\documentclass[twoside,twocolumn,9pt]{article}
\usepackage{extsizes}
\usepackage[super,sort&compress,comma]{natbib} 
\usepackage[version=3]{mhchem}
\usepackage[left=1.5cm, right=1.5cm, top=1.785cm, bottom=2.0cm]{geometry}
\usepackage{balance}
\usepackage{mathptmx}
\usepackage{sectsty}
\usepackage{graphicx} 
\usepackage{lastpage}
\usepackage[format=plain,justification=justified,singlelinecheck=false,font={stretch=1.125,small,sf},labelfont=bf,labelsep=space]{caption}
\usepackage{booktabs}
\usepackage{float}
\usepackage{fancyhdr}
\usepackage{fnpos}
\usepackage[english]{babel}
\addto{\captionsenglish}{%
  \renewcommand{\refname}{Notes and references}
}
\usepackage{array}
\usepackage{droidsans}
\usepackage{charter}
\usepackage[T1]{fontenc}
\usepackage[usenames,dvipsnames]{xcolor}
\usepackage{setspace}
\usepackage[compact]{titlesec}
\usepackage{hyperref}
\usepackage{amsmath,amssymb,amsfonts}%
\usepackage{siunitx}
\usepackage{longtable}

\usepackage{epstopdf}

\definecolor{cream}{RGB}{222,217,201}

\begin{document}

\pagestyle{fancy}
\thispagestyle{plain}
\fancypagestyle{plain}{
\renewcommand{\headrulewidth}{0pt}
}

\makeFNbottom
\makeatletter
\renewcommand\LARGE{\@setfontsize\LARGE{15pt}{17}}
\renewcommand\Large{\@setfontsize\Large{12pt}{14}}
\renewcommand\large{\@setfontsize\large{10pt}{12}}
\renewcommand\footnotesize{\@setfontsize\footnotesize{7pt}{10}}
\makeatother

\renewcommand{\thefootnote}{\fnsymbol{footnote}}
\renewcommand\footnoterule{\vspace*{1pt}%
\color{cream}\hrule width 3.5in height 0.4pt \color{black}\vspace*{5pt}} 
\setcounter{secnumdepth}{5}

\makeatletter 
\renewcommand\@biblabel[1]{#1}            
\renewcommand\@makefntext[1]%
{\noindent\makebox[0pt][r]{\@thefnmark\,}#1}
\makeatother 
\renewcommand{\figurename}{\small{Fig.}~}
\sectionfont{\sffamily\Large}
\subsectionfont{\normalsize}
\subsubsectionfont{\bf}
\setstretch{1.125} 
\setlength{\skip\footins}{0.8cm}
\setlength{\footnotesep}{0.25cm}
\setlength{\jot}{10pt}
\titlespacing*{\section}{0pt}{4pt}{4pt}
\titlespacing*{\subsection}{0pt}{15pt}{1pt}

\fancyfoot{}
\fancyfoot[LO,RE]{\vspace{-7.1pt}\includegraphics[height=9pt]{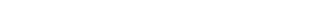}}
\fancyfoot[CO]{\vspace{-7.1pt}\hspace{13.2cm}\includegraphics{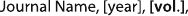}}
\fancyfoot[CE]{\vspace{-7.2pt}\hspace{-14.2cm}\includegraphics{head_foot/RF}}
\fancyfoot[RO]{\footnotesize{\sffamily{1--\pageref{LastPage} ~\textbar  \hspace{2pt}\thepage}}}
\fancyfoot[LE]{\footnotesize{\sffamily{\thepage~\textbar\hspace{3.45cm} 1--\pageref{LastPage}}}}
\fancyhead{}
\renewcommand{\headrulewidth}{0pt} 
\renewcommand{\footrulewidth}{0pt}
\setlength{\arrayrulewidth}{1pt}
\setlength{\columnsep}{6.5mm}
\setlength\bibsep{1pt}

\makeatletter 
\newlength{\figrulesep} 
\setlength{\figrulesep}{0.5\textfloatsep} 

\newcommand{\topfigrule}{\vspace*{-1pt}%
\noindent{\color{cream}\rule[-\figrulesep]{\columnwidth}{1.5pt}} }

\newcommand{\botfigrule}{\vspace*{-2pt}%
\noindent{\color{cream}\rule[\figrulesep]{\columnwidth}{1.5pt}} }

\newcommand{\dblfigrule}{\vspace*{-1pt}%
\noindent{\color{cream}\rule[-\figrulesep]{\textwidth}{1.5pt}} }

\makeatother

\twocolumn[
  \begin{@twocolumnfalse}
{\includegraphics[height=30pt]{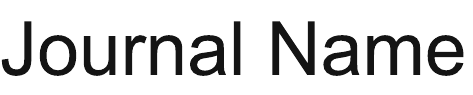}\hfill\raisebox{0pt}[0pt][0pt]{\includegraphics[height=55pt]{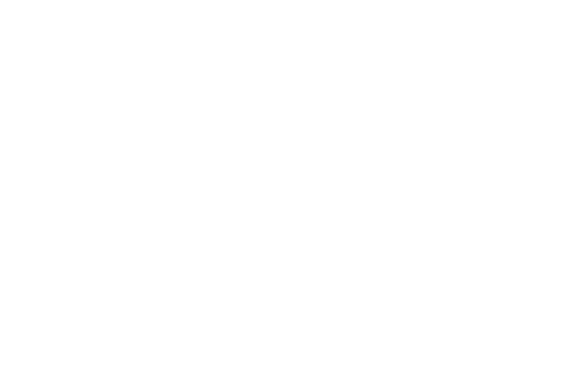}}\\[1ex]
\includegraphics[width=18.5cm]{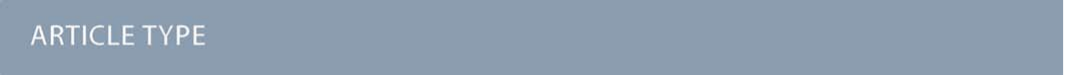}}\par
\vspace{1em}
\sffamily
\begin{tabular}{m{4.5cm} p{13.5cm} }

\includegraphics{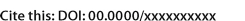} & \noindent\LARGE{\textbf{Uncertainty-Aware Continual Learning for 2D XRD Interpretation Using Simulated and Experimental Data}} \\
\vspace{0.3cm} & \vspace{0.3cm} \\

& \noindent\large{
Zihan Wang\textit{$^{a}$},
Hyun Sang Park\textit{$^{b,c}$},
Ali Rachidi\textit{$^{c}$},
David Elbert\textit{$^{c}$},
Todd C. Hufnagel\textit{$^{b,c}$},
and Wei Chen$^{\ast}$\textit{$^{a}$}
} \\

\includegraphics{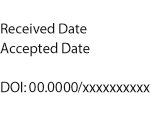} & \noindent\normalsize{Autonomous experimentation systems require characterization models that can interpret heterogeneous measurements and maintain predictive performance as data accumulate. Developing such models involves addressing domain mismatch between data sources, overconfident predictions on unfamiliar inputs, and catastrophic forgetting during sequential updates. In this work, we present a proof-of-concept framework for uncertainty-aware continual interpretation of two-dimensional X-ray diffraction (XRD) data. The framework integrates established components: maximum mean discrepancy-based domain adaptation to reduce simulation-to-experiment discrepancies, a convolutional neural network with a spectral-normalized neural Gaussian process (CNN--SNGP) for predictive uncertainty estimation, and replay with knowledge distillation for sequential model updating. We evaluate the framework using simulated and previously-collected experimental data for crystal structure classification of single-phase body-centered cubic, face-centered cubic, and hexagonal close-packed patterns. Data are organized into predefined sequential blocks to assess adaptation and retention of previously learned knowledge. Within this setting, the results show accurate classification across simulated and experimental data, improved alignment between the two domains, and reduced forgetting compared with naïve fine-tuning. Predictive uncertainty is also evaluated on unfamiliar and noisy inputs. The contribution is the integration and evaluation of these capabilities for sequential XRD interpretation. The framework provides a basis for future integration into autonomous experimentation systems, with experimental planning, instrument control, and closed-loop deployment remaining areas for future development.}
 \\

\end{tabular}

 \end{@twocolumnfalse} \vspace{0.6cm}

  ]

\renewcommand*\rmdefault{bch}\normalfont\upshape
\rmfamily
\section*{}
\vspace{-1cm}


\footnotetext{\textit{$^{a}$~Department of Mechanical Engineering,
Northwestern University, 2145 Sheridan Road, Evanston,
Illinois 60076, USA.}}

\footnotetext{\textit{$^{b}$~Department of Materials Science and Engineering,
Johns Hopkins University, 3400 N. Charles Street,
Baltimore, Maryland 21218, USA.}}

\footnotetext{\textit{$^{c}$~Hopkins Extreme Materials Institute,
Johns Hopkins University, 3400 N. Charles Street,
Baltimore, Maryland 21218, USA.}}

\footnotetext{$^{\ast}$~Corresponding author.
E-mail: weichen@northwestern.edu}




\section{Introduction}\label{sec1}

Autonomous discovery has emerged as a powerful paradigm for accelerating materials research by integrating automation, data science, and closed-loop experimentation \cite{coley2020autonomous,maffettone2023missing}. Enabled by advances in robotics, control systems, high-throughput instrumentation, and data infrastructure, the concept of the autonomous laboratory has evolved into practical autonomous experimentation systems (AES) \cite{szymanski2023autonomous,xie2023toward,kusneagent,ishizuki2023autonomous,stach2021autonomous,ferguson2022data}. These platforms typically couple automated experiment execution, characterization, data analysis, and machine-learning-guided decision-making within iterative workflows, where observations from one cycle guide subsequent experiments \cite{lookman2026materials,bukkapatnam2023autonomous}. Bayesian optimization (BO) and related active-learning strategies are established approaches for incorporating new observations, updating predictive models, and using uncertainty to guide experiment selection \cite{kusne2020fly,ament2021autonomous,wang2023bayesian}. Depending on the platform, computational simulations or theoretical predictions may also support experimental planning and improve search efficiency \cite{liang2025real}. These closed-loop approaches have demonstrated promise in composition optimization \cite{szymanski2021toward,macleod2020self,kusne2020fly,ament2021autonomous}, catalyst screening \cite{suvarna2024active,wang2023bayesian}, and microstructure design \cite{ferguson2022data,bukkapatnam2023autonomous}.

Within these systems, characterization-analysis modules interpret acquired measurements and provide information for experimental decision-making. Their effectiveness depends on accurate characterization of material structure, composition, and properties \cite{stach2021autonomous,szymanski2023autonomous,fare2022multi}. Modern characterization techniques provide complementary information across multiple spatial and temporal scales \cite{zhang2008materials,leng2013materials}; however, automated interpretation remains challenging because measurements are affected by material variations, processing conditions, and measurement noise. Analysis procedures that require extensive expert judgment and manual inspection can limit throughput and make it difficult to process the large datasets generated by high-throughput and in situ experiments. These challenges motivate computational models that can interpret characterization data and provide timely feedback to an experimental planner \cite{kusne2020fly}.

Within this workflow, experimental planning and
measurement interpretation involve complementary learning tasks.
Bayesian Optimization (BO) updates a predictive surrogate as observations accumulate and
uses uncertainty to guide subsequent experiment selection
\cite{zhang2020bayesian,wang2026arco,liu2026pareto}.
When measurement interpretation is performed by a separate
characterization model, its performance must also be maintained
as materials and measurement conditions change.
Updating the planning surrogate does not, by itself, specify how
this interpretation model should incorporate new data, assess
predictive confidence, or retain previously learned capabilities.
These requirements motivate adaptable characterization models
that provide both predictions and information about their reliability.

Deep learning (DL) provides a promising approach to automated materials characterization because it can learn complex representations directly from high-dimensional experimental data. Here, DL is considered as one option for a characterization-analysis module that can complement an established experimental planner. DL models have been used to classify crystalline phases \cite{salgado2023automated}, predict materials properties \cite{kim2021deep}, and uncover structure--function relationships that are difficult to resolve through manual analysis alone \cite{routh2021latent,kalinin2021toward}. These models can also combine information from computational and experimental datasets through knowledge transfer across different levels of approximation and data sources \cite{chen2021learning,jha2019enhancing,palizhati2022agents}. For neural-network-based characterization models operating within AES, three challenges are particularly relevant \cite{ghosh2021ensemble,kalinin2023machine}:

\begin{itemize}
\item \textbf{First, heterogeneous data sources can introduce domain mismatch.} Simulated and experimental measurements, or measurements acquired under different instrument settings and material conditions, can have different input distributions. Combining these data without accounting for their differences can limit transferability and reduce predictive accuracy.

\item \textbf{Second, predictive confidence can become unreliable on unfamiliar inputs.} Standard discriminative neural networks can assign high confidence to incorrect predictions when measurements differ from their training data. Uncertainty estimation in an experiment-selection surrogate does not automatically establish the reliability of a separate characterization model. Assessing uncertainty in measurement interpretation is therefore necessary to identify predictions that warrant additional scrutiny before informing subsequent experimental decisions.

\item \textbf{Third, sequential model updating can degrade previously learned predictive capabilities.} As new materials and measurement conditions are introduced, characterization models may need to adapt to changing data distributions. Data-management systems \cite{HTMDEC4:online} can store and organize the resulting measurements, but maintaining model performance requires an appropriate updating strategy. Retraining on the complete accumulated dataset can become increasingly costly, whereas fine-tuning only on newly acquired data can cause catastrophic forgetting. The challenge is to incorporate new evidence while retaining predictive performance on previously encountered data, particularly when the amount of historical data revisited during each update is limited.
\end{itemize}

Several research directions have been developed to address these challenges. Domain adaptation methods aim to reduce distribution mismatch between a source domain and a target domain by learning feature representations that are transferable across different input distributions. Here, a domain refers to a dataset with a distinct input distribution but the same prediction task and output space. Domain adaptation may be supervised, semi-supervised, or unsupervised, depending on the availability of target-domain labels~\cite{wilson2020survey,guan2021domain}. For example, Hu et al.~\cite{hu2024realistic} used materials from densely populated regions of the composition space as the source domain and materials from sparse or outlying regions as the target domain. They showed that incorporating the compositions of the target materials during training can improve out-of-distribution band-gap prediction and glass-forming classification. Laghane et al.~\cite{laghane2026machine} reviewed domain adaptation in electrocatalyst evaluation, including knowledge transfer across different material families, experimental conditions, and computational and experimental datasets. Li et al.~\cite{li2025cross} investigated cross-material battery health estimation, in which degradation data from one lithium-ion battery material system were used as the source domain and data from a different battery material system formed the target domain. These studies show that domain adaptation can improve transferability across heterogeneous materials datasets.

However, improved domain alignment alone does not guarantee reliable confidence estimates under unfamiliar or shifted conditions \cite{park2020calibrated}. Residual mismatch may remain after adaptation, and newly acquired data may still fall outside the learned distribution. Therefore, autonomous analysis modules should not only improve prediction accuracy across data domains but also indicate when their predictions may be unreliable. This has motivated uncertainty-aware neural networks that estimate predictive uncertainty together with class or property predictions. Among these methods, spectral-normalized neural Gaussian process (SNGP) models combine spectral normalization with a Gaussian-process output layer to produce distance-aware uncertainty estimates \cite{liu2023simple}. Such estimates are useful for autonomous systems because they can help identify unfamiliar inputs, reduce overconfident decisions, and support more reliable downstream experimental planning.

The sequential updating of characterization models within autonomous experimentation further motivates continual learning (CL) \cite{wang2024comprehensive,qu2021recent}, a paradigm in which models acquire new knowledge over time while preserving previously learned information. The focus here is on retaining previously learned predictive capabilities as data arrive sequentially and the underlying distribution changes, particularly when only a limited subset of historical data is revisited during each update. In such sequential training scenarios, conventional neural networks are prone to catastrophic forgetting, where performance on previously learned data deteriorates as the model is updated using new information. This phenomenon was first systematically documented in \cite{mccloskey1989catastrophic} and later studied extensively in connectionist models, where interference between old and new representations was identified as a major cause of forgetting \cite{french1999catastrophic,ratcliff1990connectionist}. To address this problem, a wide range of continual learning methods has been developed. These approaches are commonly grouped into three categories \cite{de2021continual,kann2023evaluation}: replay-based methods \cite{rolnick2019experience,maracani2021recall,shin2017continual,kumari2022retrospective}, which revisit past data or synthetic approximations during training; regularization-based methods \cite{zhao2024statistical,li2023fixed,kirkpatrick2017overcoming}, which restrict parameter updates to preserve prior knowledge; and parameter-isolation methods \cite{zhang2023continual,wang2023rehearsal,gao2023unified,lu2024revisiting,cai2024dynamic}, which dedicate different subsets of parameters to different tasks. Recent perspectives discuss continual learning and self-evolution as possible capabilities of intelligent scientific laboratories \cite{zhang2025position}. These proposed capabilities should be distinguished from the sequential learning and model updating already established in BO-based AES. Emerging examples in materials characterization, including defect identification from Bragg coherent diffraction imaging \cite{yildiz2024automated} and optical microscopy-based 2D material identification \cite{pandey2025cliff}, suggest its potential for enabling models to learn from new experimental data while retaining prior knowledge.

\begin{figure*}[t]
    \centering
    \includegraphics[width=0.7\textwidth]{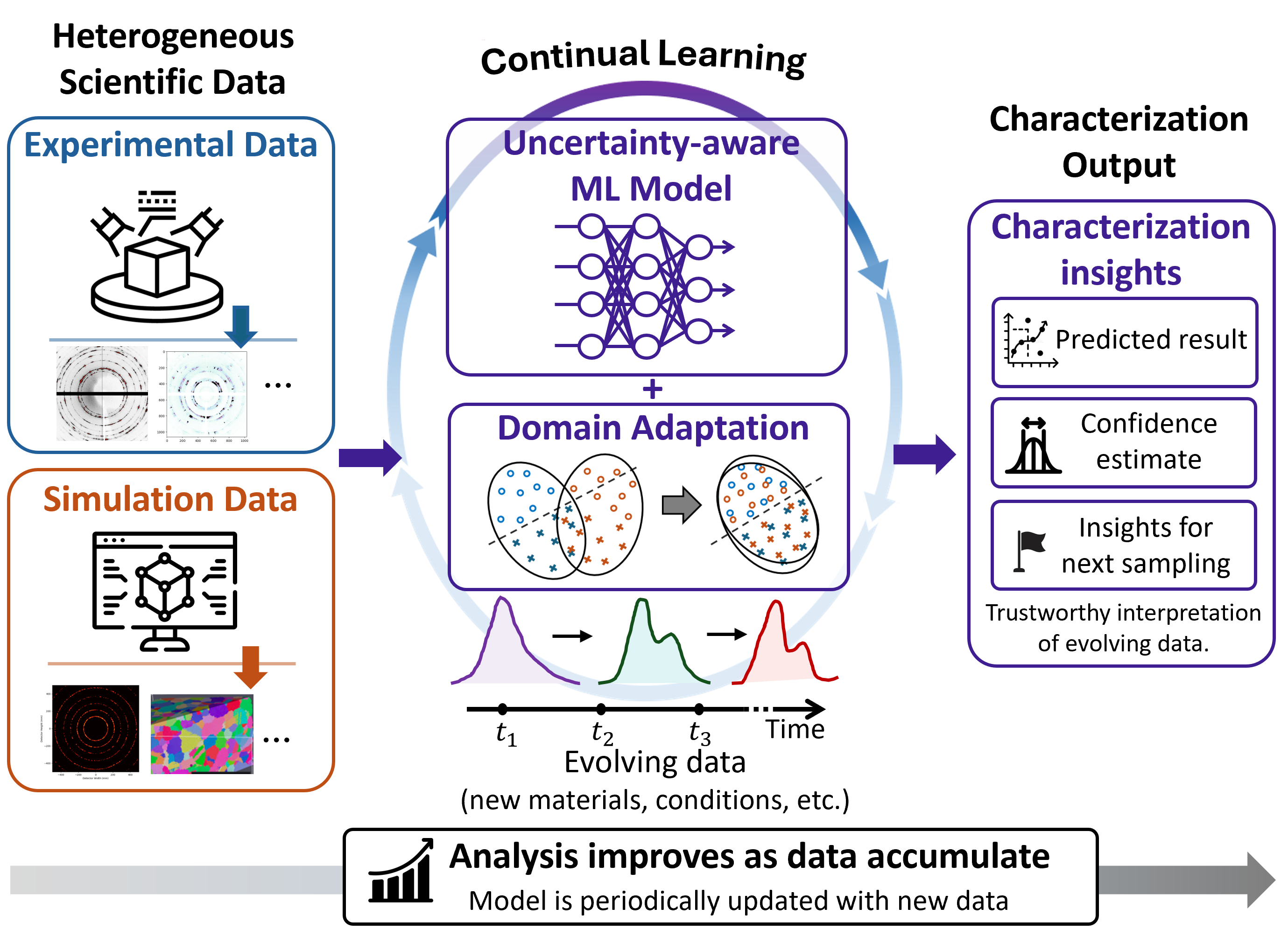}
    \caption{Conceptual framework for uncertainty-aware continual learning from heterogeneous scientific data. The framework integrates predictive uncertainty estimation, domain adaptation, and sequential model updating. This study evaluates an offline implementation using simulated and experimental 2D XRD data; applications to other characterization tasks and integration with experimental control remain future directions.}
    \label{fig:frame_all}
\end{figure*}

The objective of this work is to develop and evaluate an uncertainty-aware, continually updatable framework for sequential interpretation of simulated and experimental 2D X-ray diffraction (XRD) data, as illustrated in Figure~\ref{fig:frame_all}. The framework is evaluated as an offline proof-of-concept using predefined data blocks, with potential future application as a characterization-analysis component in autonomous experimentation systems. The present study does not select subsequent experiments or control instrument settings. Its detailed formulation and training strategy are presented in Section~\ref{sec_framework}. The evaluation focuses on crystal-structure classification of single-phase, high-symmetry body-centered cubic (BCC), face-centered cubic (FCC), and hexagonal close-packed (HCP) patterns. Specifically, a convolutional neural network-based spectral-normalized neural Gaussian process (CNN--SNGP) classifier is used to provide crystal-structure predictions together with predictive uncertainty. Simulation-to-experiment domain adaptation is incorporated to reduce discrepancies between simulated and experimental diffraction data, while a continual-learning strategy enables the model to incorporate sequentially acquired data without repeatedly retraining on the complete accumulated dataset. Together, these components provide a framework that addresses both cross-domain learning and sequential model updating. Within this offline setting, evaluation on sequential 2D XRD datasets containing simulated and experimental measurements demonstrates strong classification performance, improved simulation-to-experiment alignment, improved uncertainty behavior on unfamiliar inputs, and improved retention of previously learned knowledge during sequential updating.
The methodological contribution lies in the integration and evaluation of these established learning components. The framework is modular, with SNGP and maximum mean discrepancy (MMD) alignment serving as working choices for uncertainty estimation and domain adaptation, respectively.
The main contributions of this work are as follows:
\begin{itemize}
    \item A deep learning-based 2D XRD characterization framework is established to analyze high-dimensional diffraction images without reducing them to 1D profiles.

    \item An uncertainty-estimation component is integrated so that the proposed framework provides both class predictions and confidence information for unfamiliar, shifted, or noisy inputs. 
    
    \item A simulation-guided experimental prediction strategy is integrated into the framework to address the scarcity of labeled experimental XRD data. By aligning simulated and experimental data in a shared feature space, the framework improves the transfer of simulation-trained knowledge to real measurements.
    
    \item A continually updatable characterization-analysis module is introduced for sequential XRD data streams. The continual-learning strategy incorporates newly available data while preserving prior knowledge, reducing training time relative to joint retraining with the same backbone in the evaluated setting.

\end{itemize}
More broadly, this work provides an offline proof-of-concept for positioning characterization analysis as an adaptive, uncertainty-aware, and continually updatable component of autonomous experimentation. With appropriate integration and validation, such a module could continuously incorporate newly generated data, assess the reliability of model predictions, and adapt as experimental conditions evolve. Its potential use in closed-loop workflows would require validation of model-update triggers, uncertainty-based decision rules, and inference and retraining latency. Real-time experimental feedback and instrument control remain future extensions.

The remainder of this article is organized as follows. Section~\ref{sec_data} introduces the experimental and simulation setup, including data generation, collection, streaming, and curation. Section~\ref{sec_framework} describes the proposed deep-learning framework, including the construction of sequential learning datasets, the dual-level workflow, and the training and evaluation procedures. Section~\ref{sec_result} presents the results, with emphasis on predictive performance, simulation--experiment alignment, uncertainty estimation, and continual-learning behavior. Section~\ref{sec_discussion} discusses the scope, limitations, and broader implications of the proposed method. Finally, Section~\ref{sec_conclusion} concludes this work.

\section{Experimental and Simulation Data Pipeline}\label{sec_data}

This work considers crystal-structure classification from 2D XRD patterns as a representative example. In this setting, the input to the ML model is a measured or simulated 2D XRD diffraction pattern, and the output is the corresponding crystal-structure label (BCC, FCC, or HCP). Although the following sections focus on this specific task, the overall framework is not limited to XRD or crystal-structure classification. It is designed as a general workflow for autonomous characterization, in which continuously acquired experimental and computational data are integrated to train uncertainty-aware models for reliable prediction and decision-making.

\subsection{Simulation setup}\label{subsec:simu_setup}

The simulated dataset was constructed to represent three common high-symmetry crystal structures: BCC, FCC and HCP. These structures were selected because they are widely observed in metallic materials and produce relatively distinct diffraction signatures, making them suitable for a proof-of-concept study of 2D crystal-structure classification. To keep the initial classification problem well controlled, only single-phase materials were simulated; multiphase systems are left for future work.

Simulated 2D XRD patterns were generated using the open-source package \texttt{xrd\_simulator}~\cite{Henningsson2023}. Retained environment records support version~0.4.1, although runtime versions were not recorded in individual output files. In this package, a synthetic polycrystalline microstructure is represented by a tetrahedral mesh generated using \texttt{meshpy}. Each tetrahedral element is assigned an independent crystal lattice with a randomly sampled orientation, allowing the simulated diffraction image to capture grain-to-grain orientation variation. The intensity and size of diffraction spots are determined from crystallographic and geometric factors, including the structure factor, Lorentz factor, polarization factor, and grain-size effects. Crystallographic information, including lattice parameters, crystal structure, and Wyckoff positions, was obtained from \texttt{.cif} files downloaded on September 12, 2025 from the FIZ Karlsruhe Inorganic Crystal Structure Database. ICSD record numbers are listed in Table~\ref{tab:ICSD_reference}.

The materials included in the simulated dataset are listed in Table~\ref{tab:materials_combined}. The selected materials span the three target crystal-structure classes and provide multiple material identities within each class. The materials were selected to broaden the sampled lattice-parameter range, without considering phase stability or prevalence. Accordingly, BCC Zr (ICSD~52544) was included as a classification example, rather than as an ambient-equilibrium phase.

Microstructures were generated with a bounding radius of 70 micrometers and max cell radii of values between 5 micrometers and 45 micrometers, inclusive, in increments of 0.8 micrometers. This increment was chosen to generate a wide range of microstructures with a consistent distribution between sparse and well-populated diffraction patterns. This generated XRD patterns with progressively spottier rings. Crystallographic information in the form of lattice parameters, crystallographic structure, and Wyckoff sites were imported from cif files downloaded from the Inorganic Crystal Structure Database. Grains are generated with random orientations. The script uses \texttt{np.random.seed(None)}; neither the random state nor the sampled orientation matrices were retained, preventing exact regeneration of the original orientation realizations. Lorentz, polarization, and structure factor calculations are included in calculating the intensity of each peak.

Patterns were generated with a sample-to-detector distance of 70 mm, a detector size of 1024 by 1024 pixels using 75 micrometer pixels, and a monochromatic wavelength of 0.514 Angstroms. These are chosen to roughly match the detector and X-ray configuration of the experimental dataset, explained in Section~\ref{subsec:exp_setup}. The simulated geometry was selected to approximate, rather than exactly replicate, the experimental configuration. In particular, no pixel-wise registration between the simulated and experimental diffraction patterns was imposed. The remaining geometric and image-level differences were retained as part of the simulation-to-experiment domain discrepancy considered in this study.

\begin{table}[htbp]
\small
\caption{\textbf{Simulated and experimental materials used in this study}}
\label{tab:materials_combined}
\begin{tabular*}{\columnwidth}
{@{\extracolsep{\fill}}p{0.25\columnwidth}p{0.34\columnwidth}p{0.33\columnwidth}}
\hline
\textbf{Crystal structure}
& \textbf{Simulated materials}
& \textbf{Experimental materials} \\
\hline
FCC
& Ag, Al, Cu, Ni, Pb, Pt
& Al, Cu, Ni \\

BCC
& Mo, Nb, Fe, Ta, V, Zr
& Mo, Nb, Fe ($\alpha$-Fe) \\

HCP
& Be, Hf, Mg, Sc, Ti, Zn
& Ti, Zn \\

4H single crystal
& --
& SiC \\

4H polycrystalline
& --
& SiC \\
\hline
\end{tabular*}
\end{table}

\subsection{Experiment setup}\label{subsec:exp_setup}

XRD measurements were collected using MAXIMA, a high-throughput XRD instrument at the Artificial Intelligence for Materials Design Laboratory (AIMD-L), an autonomous laboratory at Johns Hopkins University~\cite{ParkMaxima,hufnagel_aimd-l_2026}. The scans were performed in transmission geometry using an accelerating voltage of \qty{160}{\kilo\volt}, a current of \qty{4.375}{\milli\ampere}, and an X-ray spot size of approximately \qty{250}{\micro\meter}. The sample-to-detector distance was \qty{84}{\milli\meter}, as determined from a \texttt{.poni} calibration using an alumina calibration standard. The original experimental detector images have
dimensions of $1062 \times 1028$ pixels.

These parameters were used to scan a range of available materials. Materials were chosen based on availability in a form factor conducive to transmission X-ray diffraction. The materials chosen are listed in Table \ref{tab:materials_combined}.
In addition to BCC, FCC, and HCP materials, single and polycrystalline silicon carbide are included in the scans. The SiC measurements are reserved for confidence-based detection of inputs outside the BCC/FCC/HCP structure-label set.

Each material was scanned 100 times (50 at 1 s and 50 at 3 s). Shorter scan times were chosen to decrease the signal-to-noise ratio, creating spotty diffraction rings that are harder for conventional approaches to analyze. The two exposure times were intentionally retained within the experimental
dataset to represent variation in diffraction-pattern quality and signal-to-noise ratio rather than treated as separate domains. Together with sample-dependent background, texture, grain size, and detector-related artifacts, these variations contribute to the experimental variability that is not fully represented by the simulated patterns.

Compared to simulated data, variations in the diffraction patterns from sample-to-sample exist due to influences such as the thickness of the sample, signal-to-noise ratio, pattern quality, and angular broadening. These effects are further described in \cite{ParkMaxima}. The experimental distributions of texture, grain size, and diffraction-ring spottiness are not fully represented by the simulated patterns. Because there is no data reduction to a 1D azimuthal lineout, this information is retained in the 2D image. 

Autonomous integration of experimental data with the autonomous characterization-analysis module is enabled by the event-driven architecture of AIMD-L \cite{hufnagel_aimd-l_2026}. Data are streamed from MAXIMA using OpenMSIStream \cite{Eminizer2023} to the AIMD-L data platform \cite{HTMDEC4:online}, where automated processing and curation are orchestrated. As detailed elsewhere \cite{hufnagel_aimd-l_2026}, this architecture supports sequential model-updating workflows by allowing autocurated data to be accessed programmatically through the data platform API, filtered and aggregated through orchestrated workflows, and delivered for model training \cite{ParkMaxima}.

\subsection{XRD image preprocessing}

Both simulated and experimental XRD images were converted to a common
\texttt{tiff} input format before model training.
For each original image, pixel intensities
above the image-specific 99.9th percentile were clipped to that
percentile value. Each clipped image was then independently min--max
normalized and downsampled to $256 \times 256$ pixels.
This resolution was chosen to reduce computational cost while preserving
the overall diffraction-ring geometry across the detector. The suitability
of the $256 \times 256$ resolution was further supported by a
resolution-convergence analysis based on radial peak-position preservation,
as detailed in Appendix~\ref{app:resolution_convergence}.

The preprocessed images retained a two-dimensional detector-coordinate
representation, without azimuthal integration or remapping to a common
$q$ or $2\theta$ coordinate system. Differences associated with the
simulated and experimental acquisition geometries therefore remained
part of the domain discrepancy evaluated in this study. This
representation allows the framework to process image-based inputs
while retaining the spatial organization of diffraction features.
Geometric remapping could also be incorporated while maintaining
a two-dimensional radial--azimuthal representation.

\section{Proposed Framework and Evaluation Strategy}\label{sec_framework}

\subsection{Autonomous Characterization Analysis Workflow}\label{subsec2}

As introduced in Figure~\ref{fig:frame_all}, the proposed characterization-analysis framework integrates heterogeneous data sources with an uncertainty-aware model that is updated as new data become available. The methodology is organized into two levels. The \emph{inner level} addresses cross-domain learning and predictive uncertainty by combining domain adaptation with an uncertainty-aware classifier. The \emph{outer level} addresses sequential data acquisition by continually updating the model while preserving previously learned knowledge. The following sections describe the formulation and training strategy for each level.

\subsubsection{Inner Level: Simulation-to-Experiment Feature Alignment with Uncertainty-Aware Classification}
\label{subsec:innerloop}

\begin{figure*}[t]
    \centering
    \includegraphics[width=0.85\textwidth]{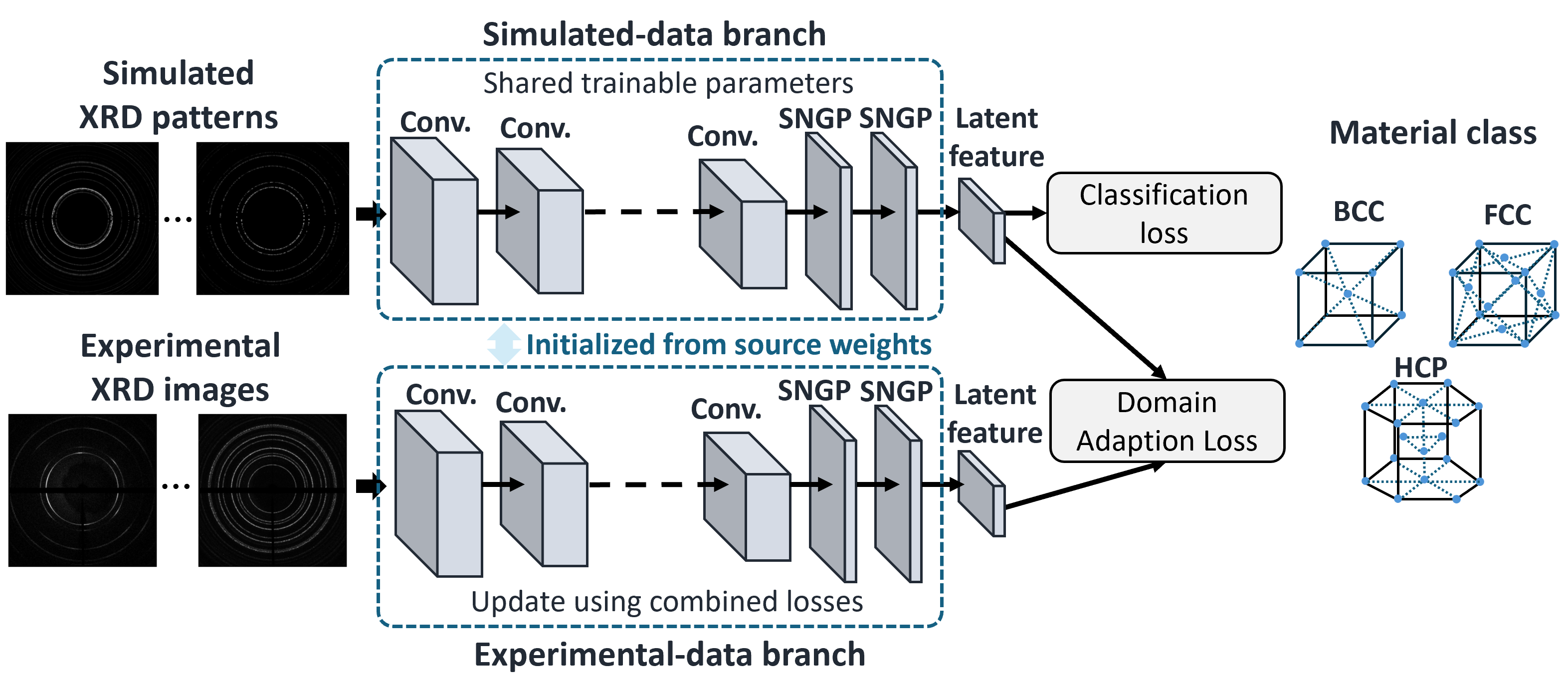}
    \caption{\textbf{Inner level: Uncertainty-aware simulation-to-experiment learning.}
    The inner level employs SNGP for uncertainty-aware classification and aligns simulated and experimental XRD feature distributions using maximum mean discrepancy. Here, Conv. denotes convolution, while BCC, FCC, and HCP denote body-centered cubic, face-centered cubic, and hexagonal close-packed, respectively.}
    \label{fig:innerloop}
\end{figure*}

Differences between simulated and experimental XRD data can reduce classification accuracy and produce unreliable predictions. To address this challenge, the inner level combines domain adaptation with a CNN--SNGP classifier to align simulated and experimental embeddings, preserve crystal-structure separation, and quantify predictive uncertainty. The overall architecture is shown in Figure~\ref{fig:innerloop}.

\begin{table*}[t]
\centering
\small
\caption{\textbf{Layers and specifications of the CNN--SNGP classifier.}
CNN denotes convolutional neural network; SNGP denotes
spectral-normalized neural Gaussian process; SN--Dense denotes
a spectrally normalized dense layer; and ReLU denotes the
rectified linear unit activation.
The CNN--SNGP--adapt model uses the same architecture with
additional domain-adaptation losses.}
\label{tab:model_architecture}
\renewcommand{\arraystretch}{1.15}
\begin{tabular*}{\textwidth}
{@{\extracolsep{\fill}}p{0.14\textwidth}p{0.25\textwidth}p{0.53\textwidth}}
\hline
\textbf{Block} & \textbf{Layer} & \textbf{Specification} \\
\hline

Feature extractor
& Input
& $1 \times 256 \times 256$ grayscale image \\

& Convolutional block 1
& \begin{tabular}[t]{@{}l@{}}
$7 \times 7$ convolution, 32 channels, stride 1, padding 3; ReLU \\
$7 \times 7$ convolution, 32 channels, stride 1, padding 3; ReLU \\
$2 \times 2$ max pooling, stride 1, no padding \\
Output: $32 \times 255 \times 255$
\end{tabular} \\

& Convolutional block 2
& \begin{tabular}[t]{@{}l@{}}
$7 \times 7$ convolution, 64 channels, stride 1, padding 3; ReLU \\
$7 \times 7$ convolution, 64 channels, stride 1, padding 3; ReLU \\
$2 \times 2$ max pooling, stride 2, no padding \\
Output: $64 \times 127 \times 127$
\end{tabular} \\

& Convolutional block 3
& \begin{tabular}[t]{@{}l@{}}
$7 \times 7$ convolution, 96 channels, stride 1, padding 3; ReLU \\
$7 \times 7$ convolution, 96 channels, stride 1, padding 3; ReLU \\
$2 \times 2$ max pooling, stride 2, no padding \\
Output: $96 \times 63 \times 63$
\end{tabular} \\

& Flatten
& $96 \times 63 \times 63 = 381{,}024$ features \\

& Dense
& $381{,}024 \rightarrow 2{,}048$; linear, without spectral normalization \\

& SN--Dense 1
& $2{,}048 \rightarrow 512$; spectral normalization, followed by ReLU \\

& SN--Dense 2
& $512 \rightarrow 256$; spectral normalization, followed by ReLU \\

\hline

GP head
& Random Fourier features
& \begin{tabular}[t]{@{}l@{}}
Fixed random Fourier mapping: $256 \rightarrow 64$ \\
RBF kernel approximation with $\gamma = 1/64$
\end{tabular} \\

& GP mean
& Linear mapping $64 \rightarrow 3$, without bias \\

& Predictive uncertainty
& \begin{tabular}[t]{@{}l@{}}
$64 \times 64$ precision matrix with ridge parameter 1 \\
Input-dependent scalar predictive variance shared across classes
\end{tabular} \\

& Class probabilities
& Mean-field-adjusted softmax over BCC, FCC, and HCP \\

\hline
\end{tabular*}
\end{table*}

\textbf{Uncertainty-aware classification via CNN--SNGP.}
To support predictive modeling while retaining uncertainty information, the proposed module uses a CNN--SNGP model as the predictive backbone. The CNN serves as a feature extractor that maps high-dimensional input data $\mathbf{x}$ into a learned latent representation, while the SNGP layer provides distance-aware predictive uncertainty in the resulting feature space. More generally, SNGP can be integrated with suitable neural-network encoders to accommodate a broad range of high-dimensional data representations. The detailed CNN--SNGP architecture adopted in this work is summarized in Table~\ref{tab:model_architecture}.

A CNN first extracts features from a 2D image data $\mathbf{x}$, producing an intermediate representation 
\begin{equation}
    \mathbf{z}_{\mathrm{cnn}} = f_{\theta}(\mathbf{x}) \in \mathbb{R}^{D_{\mathrm{cnn}}}.
\end{equation}
Here, $\theta$ denotes the trainable parameters of the CNN feature extractor $f_{\theta}$.
The extracted features are then passed through spectrally normalized fully connected layers, which constrain the largest singular value of each weight matrix. This enforces a Lipschitz-bounded mapping, helping preserve local distances in the latent space and reducing feature collapse. 

The spectrally normalized features are then projected into a random Fourier feature (RFF) space, which approximates a Gaussian kernel. These RFF embeddings are used by a GP-style final layer that models the predictive logits as a Gaussian random variable:
\begin{equation}
    \mathbf{g}(\mathbf{x})
    \sim 
    \mathcal{N}\!\big(\boldsymbol{\mu}(\mathbf{x}),\, \boldsymbol{\Sigma}(\mathbf{x})\big).
    \label{eq:sngp_gaussian}
\end{equation}
where $\boldsymbol{\mu}(\mathbf{x}) \in \mathbb{R}^{C}$ is the predictive mean vector, where $C$ is the number of classes. Each component $\mu_c(\mathbf{x})$ is the mean logit for class $c$.
$\boldsymbol{\Sigma}(\mathbf{x}) \in \mathbb{R}^{C \times C}$ is the predictive covariance matrix, capturing how uncertain the model is about the logits. In practice, SNGP uses a scalar variance
\begin{equation}
    \sigma^2(\mathbf{x})
    =
    \mathrm{Var}\!\left[g_c(\mathbf{x})\right],
    \label{eq:predictive_variance}
\end{equation}
    identical across classes, computed from a ridge-regularized precision matrix formed by additive accumulation of the random Fourier feature statistics. Here, $\sigma^2(x)$ denotes the approximate GP posterior variance in latent-logit space and equals the square of the standard deviation returned by the GP head. It provides a model-dependent proxy for epistemic uncertainty conditional on the trained encoder and fixed random Fourier features. It is distinct from variance in the predicted class probabilities.

In the classification problem considered in this work, class probabilities are required rather than the logit means directly. Because the logits are Gaussian random variables, the predictive probability for class $c$ is the expectation of a softmax under a Gaussian distribution:
\begin{equation}
    p(y=c \mid \mathbf{x})
    = 
    \mathbb{E}_{\mathbf{g}\sim \mathcal{N}(\boldsymbol{\mu}, \boldsymbol{\Sigma})}
    \!\big[\mathrm{softmax}(\mathbf{g})_c\big].
    \label{eq:gaussian_softmax_integral}
\end{equation}
For scalar logit variance $\sigma^2(\mathbf{x})$, the predictive probability \cite{lu2020mean} becomes:
\begin{equation}
    p(y=c \mid \mathbf{x})
    =
    \mathrm{softmax}\!\left(
        \frac{\boldsymbol{\mu}(\mathbf{x})}
        {\sqrt{\,1 + \frac{\pi}{8}\sigma^2(\mathbf{x})}}
    \right)_c .
    \label{eq:sngp_softmax}
\end{equation}

Equation~\ref{eq:sngp_softmax} is a deterministic mean-field approximation to the Gaussian-logit expectation in Eq.~\ref{eq:gaussian_softmax_integral}.

\textbf{Domain adaptation via Maximum Mean Discrepancy (MMD).}
A model trained only on simulated XRD patterns may not transfer directly
to experimental XRD patterns because the two domains can differ in peak
shape, noise level, background, preferred orientation, and detector
response. To reduce this simulation-to-experiment gap, we use MMD to
encourage simulated and experimental samples to occupy similar regions
in the model's latent feature space.

The classifier is trained using labeled simulated samples and a labeled
subset of experimental training samples, while unlabeled experimental
samples support distribution alignment. This constitutes semi-supervised
domain adaptation.

Let
$\mathcal{D}_s=\{(\mathbf{x}_i^s,y_i^s)\}_{i=1}^{N_s}$
denote the labeled simulated training set and
$\mathcal{D}_t=\{\mathbf{x}_j^t\}_{j=1}^{N_t}$
the experimental training set. We reveal the true labels of a selected
target subset indexed by $\mathcal{I}_a$, forming the anchor set
\[
\mathcal{D}_a
=
\{(\mathbf{x}_j^t,y_j^t):j\in\mathcal{I}_a\},
\qquad
\widehat{\rho}_a
=
\frac{|\mathcal{I}_a|}{N_t}.
\]
Here, $\widehat{\rho}_a$ is the realized target training-label fraction.
The requested fraction is $\rho_a=0.10$ for the fixed-model comparisons
and $\rho_a=0.10$ for the continual-learning comparison described in
Section~4.4. The realized fraction may differ slightly because of integer
sample counts. The remaining experimental training samples are treated
as unlabeled. Validation and test samples are excluded from anchor
selection. The anchor fraction describes training-label availability;
it is distinct from the loss weights defined below.

For each input XRD image, the CNN--SNGP feature extractor produces a
latent feature vector,
$\mathbf{z}=f_{\theta}(\mathbf{x})$,
taken from the final spectrally normalized dense layer. During
adaptation, training mini-batches contain labeled simulated samples,
unlabeled experimental samples, and, when available, labeled
experimental anchors. These provide source-domain classification
supervision, target-domain distribution information, and target-domain
classification supervision, respectively.

Let
$\mathcal{Z}_s=\{\widetilde{\mathbf{z}}_s^i\}_{i=1}^{n_s}$
and
$\mathcal{Z}_t=\{\widetilde{\mathbf{z}}_t^j\}_{j=1}^{n_t}$
denote the $\ell_2$-normalized latent features of simulated and unlabeled
experimental samples in a mini-batch. MMD measures the difference
between their feature distributions:
\begin{equation}
\begin{aligned}
\mathrm{MMD}^{2}(\mathcal{Z}_s,\mathcal{Z}_t)
&=
\mathbb{E}_{\widetilde{\mathbf{z}}_s,
             \widetilde{\mathbf{z}}'_s}
\left[
k_{\sigma}(\widetilde{\mathbf{z}}_s,
           \widetilde{\mathbf{z}}'_s)
\right]
+
\mathbb{E}_{\widetilde{\mathbf{z}}_t,
             \widetilde{\mathbf{z}}'_t}
\left[
k_{\sigma}(\widetilde{\mathbf{z}}_t,
           \widetilde{\mathbf{z}}'_t)
\right] \\
&\quad
-
2\mathbb{E}_{\widetilde{\mathbf{z}}_s,
              \widetilde{\mathbf{z}}_t}
\left[
k_{\sigma}(\widetilde{\mathbf{z}}_s,
           \widetilde{\mathbf{z}}_t)
\right],
\end{aligned}
\label{eq:mmd}
\end{equation}
where $k_{\sigma}$ is a Gaussian kernel,
\begin{equation}
k_{\sigma}(\widetilde{\mathbf{z}}_i,
           \widetilde{\mathbf{z}}_j)
=
\exp\left(
-\frac{
\|\widetilde{\mathbf{z}}_i-\widetilde{\mathbf{z}}_j\|_2^2
}{2\sigma^2}
\right).
\label{eq:mmd_kernel}
\end{equation}
Feature normalization reduces sensitivity to feature scale. We use
a single Gaussian kernel with fixed bandwidth $\sigma=1.0$.
The distribution-alignment loss is
$\mathcal{L}_{\mathrm{MMD}}
=\mathrm{MMD}^{2}(\mathcal{Z}_s,\mathcal{Z}_t)$.
A smaller value indicates closer agreement between the source and
target feature distributions, but does not by itself guarantee
class-consistent alignment.

The labeled simulated samples provide source-domain supervision through
\[
\mathcal{L}_{\mathrm{cls}}^{\mathrm{src}}
=
-\frac{1}{|\mathcal{B}_s|}
\sum_{(\mathbf{x},y)\in\mathcal{B}_s}
\log p_{\theta}(y\mid\mathbf{x}),
\]
where $\mathcal{B}_s$ is a labeled source mini-batch.
The labeled experimental anchors provide additional target-domain
supervision through
\begin{equation}
\mathcal{L}_{\mathrm{cls}}^{\mathrm{tgt}}
=
-\frac{1}{|\mathcal{B}_a|}
\sum_{(\mathbf{x},y)\in\mathcal{B}_a}
\log p_{\theta}(y\mid\mathbf{x}),
\label{eq:target_anchor_classification}
\end{equation}
where $\mathcal{B}_a$ contains labeled target anchors.
Only revealed anchor labels contribute to this target-domain
supervised loss.

To encourage class-consistent alignment, we match source and
target-anchor class prototypes in the latent feature space.
For each class represented in both mini-batches, the prototypes are
\begin{equation}
\boldsymbol{\mu}_c^s
=
\frac{1}{|\mathcal{B}_s^c|}
\sum_{(\mathbf{x},y)\in\mathcal{B}_s^c}
f_{\theta}(\mathbf{x}),
\qquad
\boldsymbol{\mu}_c^a
=
\frac{1}{|\mathcal{B}_a^c|}
\sum_{(\mathbf{x},y)\in\mathcal{B}_a^c}
f_{\theta}(\mathbf{x}),
\label{eq:class_prototypes}
\end{equation}
where $\mathcal{B}_s^c$ and $\mathcal{B}_a^c$ contain source
samples and target anchors with true class label $c$, respectively.
The prototype-alignment loss is
\begin{equation}
\mathcal{L}_{\mathrm{anchor}}
=
\frac{1}{|\mathcal{C}_a|}
\sum_{c\in\mathcal{C}_a}
\|\boldsymbol{\mu}_c^s-\boldsymbol{\mu}_c^a\|_2^2,
\label{eq:anchor_alignment}
\end{equation}
where $\mathcal{C}_a$ contains the classes represented in both
mini-batches. The target classification loss is omitted when
$\mathcal{B}_a$ is empty, and the prototype-alignment loss is
omitted when $\mathcal{C}_a$ is empty.

The combined adaptation objective is
\begin{equation}
\mathcal{L}_{\mathrm{inner}}
=
\mathcal{L}_{\mathrm{cls}}^{\mathrm{src}}
+
\lambda_{\mathrm{tgt}}
\mathcal{L}_{\mathrm{cls}}^{\mathrm{tgt}}
+
\lambda_{\mathrm{mmd}}
\mathcal{L}_{\mathrm{MMD}}
+
\lambda_{\mathrm{anchor}}
\mathcal{L}_{\mathrm{anchor}}.
\label{eq:inner_objective}
\end{equation}
Here, $\mathcal{L}_{\mathrm{cls}}^{\mathrm{src}}$ provides
simulated-data supervision,
$\mathcal{L}_{\mathrm{cls}}^{\mathrm{tgt}}$ provides labeled
experimental-anchor supervision,
$\mathcal{L}_{\mathrm{MMD}}$ aligns the overall source and
unlabeled-target feature distributions, and
$\mathcal{L}_{\mathrm{anchor}}$ aligns corresponding class prototypes.
The weights $\lambda_{\mathrm{tgt}}$, $\lambda_{\mathrm{mmd}}$,
and $\lambda_{\mathrm{anchor}}$ control the relative contributions
of these terms.

\subsubsection{Outer Level: Sequential Model Updating}
\label{subsec:outer-loop}

\begin{figure}[t]
    \centering
    \includegraphics[width=\columnwidth]{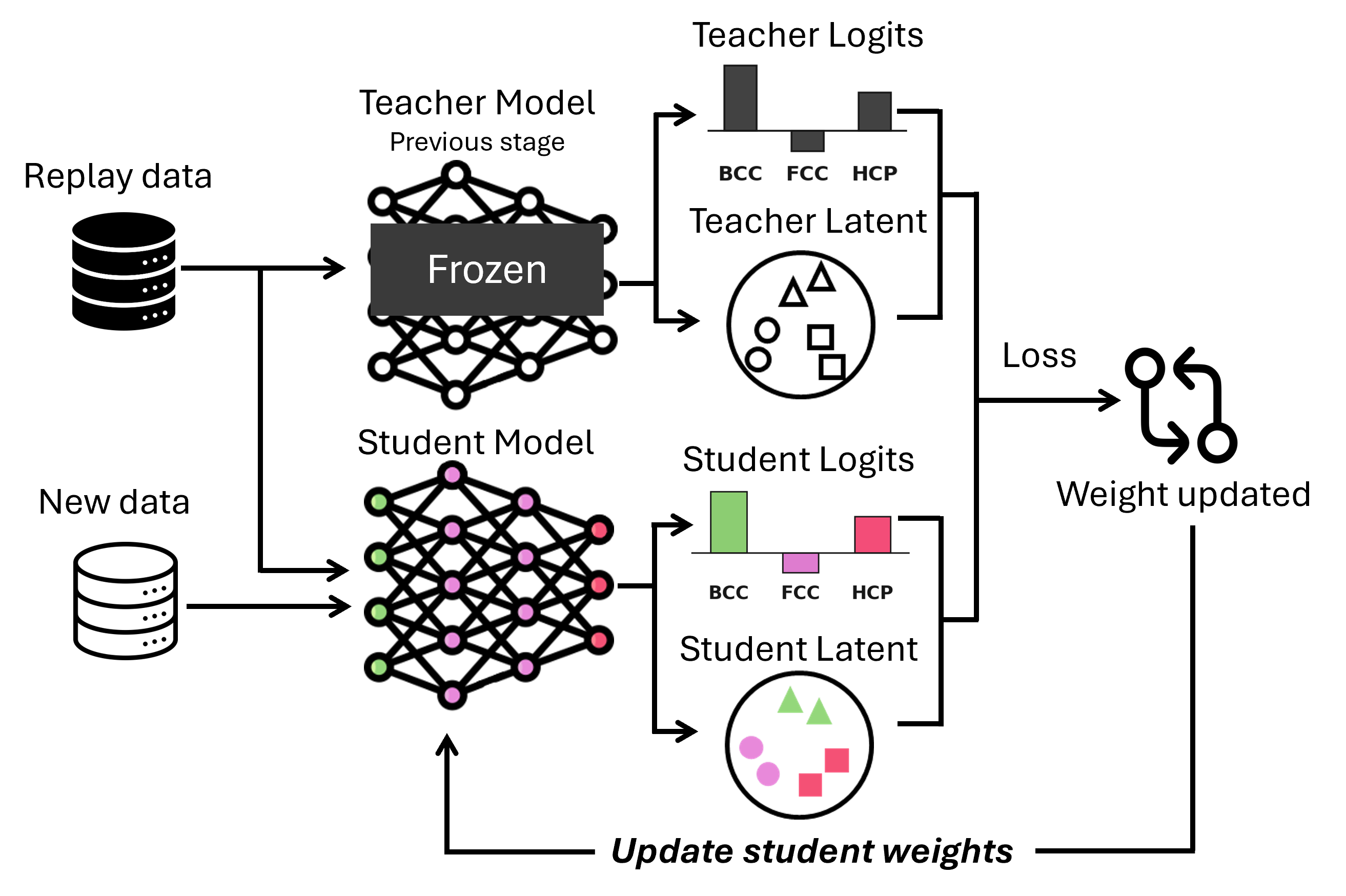}
    \caption{\textbf{Outer-level sequential model updating with replay and distillation.}
The frozen teacher from the preceding stage and the student process the
same replayed inputs. Logit distillation uses Kullback--Leibler divergence
to match the student's temperature-scaled class probabilities to the
teacher's. Feature distillation uses a normalized squared error between
their 256-dimensional pre-head features. The student is updated using
the combined objective in Eq.~(14).}
    \label{fig:overall}
\end{figure}

Sequentially acquired data may shift the underlying distribution, while naïve fine-tuning can overwrite previously learned decision boundaries and cause catastrophic forgetting. To address this challenge, the outer level (Figure~\ref{fig:overall}) updates the model using replay-based continual learning, combining each new data block with teacher--student distillation and feature-geometry alignment while optimizing the inner-level loss.

The training stream is represented as a sequence of data blocks
$\{\mathcal{D}^{(1)},\ldots,\mathcal{D}^{(N_{\mathrm{blk}})}\}$,
where each block contains simulated source data and experimental target
data. In the proposed continual-learning strategy, for $t>1$ the student
is initialized from the model selected at the end of the preceding block,
$\Theta^{(t)} \leftarrow \Theta^{(t-1)}$. A frozen copy of that model
serves as the teacher, with
$\widetilde{\Theta}^{(t)} \leftarrow \Theta^{(t-1)}$,
and provides stable targets for replay-based distillation.

To preserve knowledge from earlier blocks, we maintain a bounded replay
buffer $\mathcal{M}=\mathcal{M}_s\cup\mathcal{M}_t$. The source subset
$\mathcal{M}_s$ stores labeled simulated samples. The target subset
$\mathcal{M}_t$ stores experimental samples and retains the labels of
replayed anchors whose labels were revealed in their original training
blocks. The remaining experimental replay samples are treated as
unlabeled. From block 2 onward, each optimization step combines current-block
mini-batches with mini-batches from the available replay subsets.

We apply cross-entropy to labeled replay samples. Let
$\mathcal{B}_{d,\mathrm{lab}}^{\mathrm{rep}}$ denote the labeled samples in
the replay mini-batch for domain $d\in\{s,t\}$. This set includes all sampled
simulated scans in the source domain and the sampled experimental anchors
in the target domain. With $g_\Theta(x)$ denoting the student logits,
the replay cross-entropy is
\begin{equation}
\mathcal{L}_{\mathrm{rCE}} =
\sum_{d\in\{s,t\}}
\mathbb{E}_{(x,y)\in\mathcal{B}_{d,\mathrm{lab}}^{\mathrm{rep}}}
\!\left[\ell_{\mathrm{CE}}\!\left(g_\Theta(x),y\right)\right],
\end{equation}
where a domain contributes zero if its replay mini-batch contains no
labeled samples.

The student also matches the frozen teacher's softened class
probabilities and pre-head features on the same replay inputs $x$.
For $\Psi\in\{\Theta,\widetilde{\Theta}\}$, let
$p_\Psi^{(T)}(x)=\operatorname{softmax}(g_\Psi(x)/T)$, where
$g_\Psi(x)$ denotes the logits and $T$ is the distillation temperature.
Let $h_\Psi(x)\in\mathbb{R}^{256}$ denote the pre-head features immediately
before the random Fourier feature mapping. The two distillation losses are
\begin{equation}
\begin{aligned}
\mathcal{L}_{\mathrm{KD}}
&= T^2\left\langle D_{\mathrm{KL}}\!\left(
p_{\widetilde{\Theta}}^{(T)}(x) \,\middle\|\, p_\Theta^{(T)}(x)
\right)\right\rangle_{\mathrm{rep}},\\
\mathcal{L}_{\mathrm{feat}}
&= \frac{\left\langle
\|h_\Theta(x)-h_{\widetilde{\Theta}}(x)\|_2^2
\right\rangle_{\mathrm{rep}}}{256s_h^2}.
\end{aligned}
\end{equation}
Here, $\langle\cdot\rangle_{\mathrm{rep}}$ denotes the mean within each
nonempty replay mini-batch, followed by an equal-weight mean across
domains. The shared feature scale $s_h$ is
computed as the square root of the mean coordinate variance of the
student's pre-head features on current-block training scans and retained
replay scans, with a lower bound of $10^{-4}$. Both losses apply to labeled
and unlabeled replay samples.

The resulting outer-level objective for block $t$ is
\begin{equation}
\label{eq:outer_loss}
\begin{aligned}
\mathcal{L}_{\mathrm{outer}}^{(t)}
&= \mathcal{L}_{\mathrm{fit}}^{(t)}
   + \lambda_{\mathrm{rCE}}\mathcal{L}_{\mathrm{rCE}}\\
&\quad + \lambda_{\mathrm{KD}}\mathcal{L}_{\mathrm{KD}}
       + \lambda_{\mathrm{feat}}\mathcal{L}_{\mathrm{feat}}.
\end{aligned}
\end{equation}
The replay and distillation terms are zero in the first block. For
$t>1$, we use $T=2$, $\lambda_{\mathrm{rCE}}=1$,
$\lambda_{\mathrm{KD}}=3$, and $\lambda_{\mathrm{feat}}=1$.
The phase-specific fitting loss $\mathcal{L}_{\mathrm{fit}}^{(t)}$
includes supervised cross-entropy. Global MMD aligns the simulated and experimental batches, and
class-prototype alignment uses revealed labels and confident target
pseudo-labels. Both alignment terms use $\ell_2$-normalized pre-head
features and have weight $2$. An unlabeled target scan receives a
pseudo-label when its maximum softmax probability is at least $0.90$.
Such scans also contribute to pseudo-label cross-entropy with weight
$0.1$.

At the end of each block, the bounded replay buffer is refreshed from
the previously retained scans and the current block's training scans.
The proposed strategy selects simulated scans by class-wise herding and
experimental scans by feature diversity. The model selected for block $t$
is then carried forward as the initial student and frozen teacher for the
next block.

\subsection{Data Partitioning for Fixed and Sequential Evaluation}
\label{subsec:data_sequence}

The proposed framework must be evaluated under both cross-domain transfer and continuously evolving data. Accordingly, the simulated and experimental XRD datasets are organized into two complementary settings. The fixed setting evaluates simulation-to-experiment transfer, uncertainty estimation, and material-level generalization using all available training data, whereas the sequential setting divides the data into ordered blocks to assess continual model updating as new measurements become available. The revised evaluation separates training, validation, and test data using an 80\%/10\%/10\% split within the development data, with designated material-held-out and out-of-label-space datasets reserved for independent testing. Validation data are used for early stopping and model selection; test data are excluded from these decisions.

\textbf{Fixed-training evaluation.}
In the fixed-training setting, the model is trained using the pooled training data rather than sequential updates, with repeated training across three random initializations. This setting represents the case where the currently available simulated and experimental data are pooled before model training. It is used to evaluate whether the model can learn the known crystal-structure label space, transfer from simulated to experimental XRD patterns, and provide meaningful uncertainty estimates for unfamiliar inputs.

To test material-level generalization, we perform
leave-one-material-out evaluation across eight materials
(Al, Cu, Fe, Mo, Nb, Ni, Ti, and Zn). In each fold, the held-out
material is excluded from training and validation in both the
simulated and experimental domains and reserved for known-class
testing within the BCC/FCC/HCP label space. Three training repeats
with different random seeds are performed for each holdout,
yielding 24 fits per method, with matched data partitions,
experimental training-anchor masks, and validation information
across methods.
In addition, experimental SiC measurements are used as
out-of-label-space inputs because SiC does not belong to any of
the three trained crystal-structure classes.
Here, HCP refers to the elemental hexagonal
close-packed structure with ABAB stacking, rather than all
structures with hexagonal symmetry. Although 4H-SiC is hexagonal,
its silicon--carbon atomic basis and stacking arrangement differ
from this HCP structure. Single-crystal and polycrystalline SiC
also introduce different diffraction morphologies and orientation
distributions.
Noise-only images are also evaluated as invalid inputs.
These SiC and noise-only datasets are not assigned BCC/FCC/HCP
labels and are used only for uncertainty-based evaluation.
Table~\ref{tab:dataset_fix} summarizes the dataset partitions
used in the fixed-training evaluation.

\textbf{Sequential-update evaluation.}
The second setting evaluates continual learning under sequential data arrival. To
emulate the progressive acquisition of XRD measurements in an autonomous
characterization workflow, the simulated and experimental datasets are organized
into a sequence of four data blocks, denoted as
$\mathcal{D}^{(1)}, \ldots, \mathcal{D}^{(4)}$. Each block corresponds to one
model-update stage and contains one simulated source-domain subset,
$\mathcal{D}^{(t)}_{\mathrm{sim}}$, and one experimental target-domain subset,
$\mathcal{D}^{(t)}_{\mathrm{exp}}$:
\begin{equation}
\mathcal{D}^{(t)} =
\left\{
\mathcal{D}^{(t)}_{\mathrm{sim}},
\mathcal{D}^{(t)}_{\mathrm{exp}}
\right\}.
\end{equation}
The simulated subset provides labeled source-domain samples for supervised
crystal-structure classification, whereas the experimental subset provides
target-domain samples for simulation-to-experiment adaptation and evaluation.
Training, validation, and test subsets remain separate throughout
sequential updating, and test samples are excluded from adaptation, replay,
and model selection.

The four blocks $\mathcal{D}^{(1)}$--$\mathcal{D}^{(4)}$ are introduced one at a
time to mimic the progressive arrival of measurements in an autonomous laboratory.
The label space remains fixed throughout the sequence as BCC, FCC, and HCP, while
the data distribution changes across stages as new simulated and experimental
materials are introduced. The blocks are constructed to emulate sequential
acquisition with controlled distribution shifts: some material identities reappear
across stages using disjoint scans or independently generated simulated images,
while others are introduced only in later stages. Each scan is assigned to exactly one data block and one data split. To assess sensitivity to initialization and arrival order, each
training strategy is evaluated using three random initializations and three
block orders, yielding nine runs per strategy. Results are averaged across
orders within each initialization before calculating the mean and sample
standard deviation across the three initializations. All strategies use
matched data partitions and block orders.

\begin{figure*}[t]
    \centering
    \includegraphics[width=0.8\textwidth]{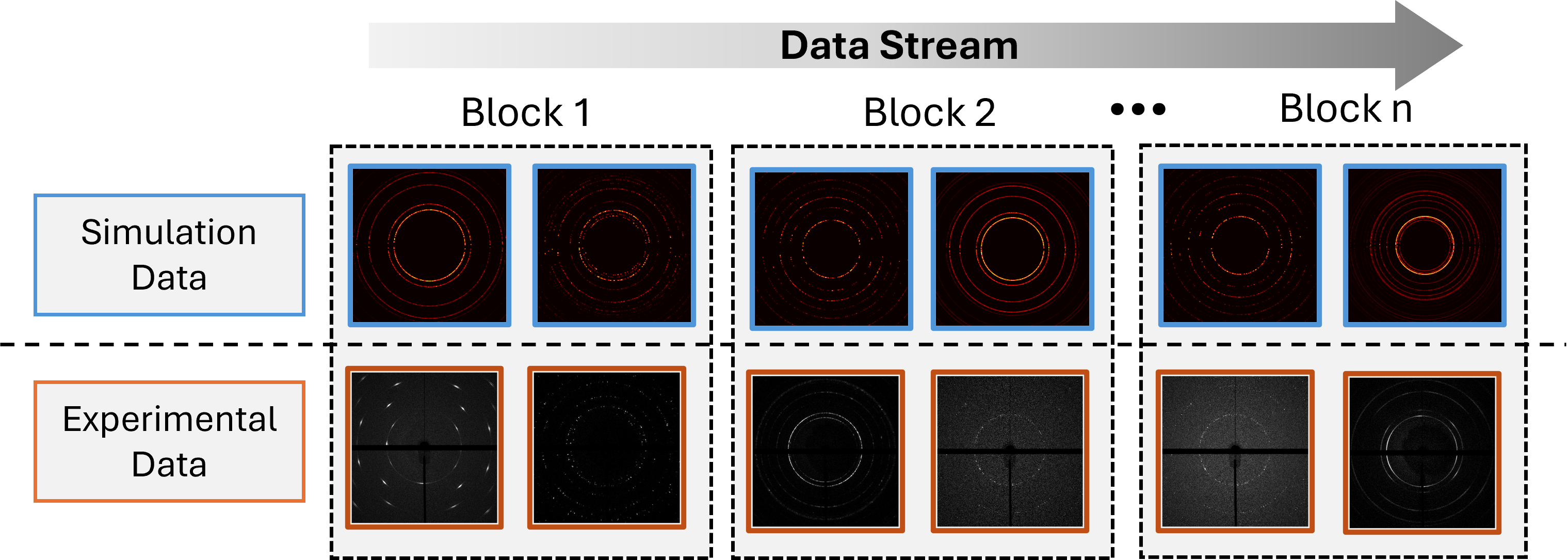}
    \caption{\textbf{Sequential organization of simulated and experimental XRD datasets.}
    Each dashed box represents one model-update stage and contains one simulated data block and one experimental data block.
    The known BCC, FCC, and HCP label space remains fixed across stages, while new material identities and experimental measurements are introduced over time to create sequential distribution shifts.
    The illustrated sequence represents one block order; three orders are evaluated in the repeated-training comparison.}
    \label{fig:data}
\end{figure*}

As illustrated in Figure~\ref{fig:data}, the simulated and experimental datasets
are paired stage by stage. Within each dashed box, simulated patterns are shown in
the top row and experimental patterns are shown in the bottom row. Changes across
blocks are reflected by differences in diffraction-ring positions, ring intensities,
spot distributions, and background characteristics, indicating variations in
material identity and measurement conditions over time. In addition, paired
simulated and experimental patterns differ within each block because experimental
measurements include noise, background variation, detector shadows, and other
instrument-related artifacts that are not fully captured in the simulated data.
The composition of the sequential blocks is summarized in
Table~\ref{tab:dataset_summary}.

\subsection{Evaluation Metrics and Training Strategy}
\label{sec:metrics}

We evaluate classification performance, predictive-probability
quality, unfamiliar-input detection, latent-space organization,
and knowledge retention during sequential updating.

\textbf{Classification and probability quality.}
Classification is assessed using accuracy, balanced accuracy
(mean class recall), and macro-F1 (mean classwise F1).
Predictions are obtained from
$\hat{y}_i=\arg\max_k p_{\theta}(y=k\mid x_i)$.
Negative log-likelihood (NLL) evaluates the probability assigned
to the true class, while the multiclass Brier score measures
squared differences between predicted probabilities and one-hot
labels. Expected calibration error (ECE) measures the
sample-weighted absolute gap between accuracy and confidence
within confidence bins. Lower NLL, Brier score, and ECE indicate
better predictive-probability quality.

\textbf{Confidence and unfamiliar-input detection.}
Predictive confidence and the OOD ranking score are defined as
\begin{equation}
C_i=\max_k p_{\theta}(y=k\mid x_i),
\qquad
u_i=1-C_i.
\end{equation}
Larger $u_i$ indicates greater unfamiliarity, but is not a
calibrated probability of novelty or a pure measure of epistemic
uncertainty. With OOD inputs treated as positive examples,
detection is assessed using AUROC, AUPR, and FPR95.
FPR95 measures the fraction of known-class inputs rejected at an
operating point attaining at least 95\% OOD detection.
OOD false acceptance is the fraction of OOD inputs accepted at
the evaluated rejection threshold.

\textbf{Latent-space organization.}
We use pre-head features to quantify class separation through
the silhouette score and source--target agreement through the
mean class-centroid distance:
\begin{equation}
S=\frac{1}{N}\sum_{i=1}^{N}
\frac{b_i-a_i}{\max(a_i,b_i)},
\qquad
D_{\mathrm{align}}
=\frac{1}{|\mathcal{K}|}
\sum_{k\in\mathcal{K}}
\|\boldsymbol{\mu}^{\mathrm{sim}}_k
-\boldsymbol{\mu}^{\mathrm{exp}}_k\|_2.
\end{equation}
Here, $a_i$ is the mean within-class distance, $b_i$ is the
smallest mean distance to another class, and $\mathcal{K}$
contains classes represented in both domains. Higher silhouette
and lower centroid distance indicate stronger class separation
and closer class-conditional alignment.
The fixed-model comparison uses jointly standardized simulated
and experimental validation features for both metrics.
For continual learning, $S_{\mathrm{raw}}$ uses raw features,
whereas $D_{\mathrm{align,std}}$ uses coordinate-standardized
validation features. Standardization uses
$\tilde{z}_{i,m}=(z_{i,m}-\bar{z}_m)/(\sigma_m+10^{-8})$.
Neither metric is calculated from t-SNE coordinates.

\textbf{Continual-learning performance.}
Scan-pooled accuracy is complemented by average accuracy (AA),
which weights introduced blocks equally. Signed forgetting
compares final accuracy with earlier peak performance;
backward transfer compares final performance with performance
when a block was first learned. Forward transfer compares
pre-introduction performance with a random-initialization
baseline. Old- and new-block accuracies separately assess
retention and adaptation. Training time includes validation, GP precision reconstruction, and retries.

\textbf{Training and reporting.}
Compared methods receive matched data partitions, experimental
training-anchor masks, and validation information.
Model selection and early stopping use validation data;
test data are reserved for assessment. Results are reported
as mean $\pm$ sample standard deviation, with fold pooling
and block-order aggregation specified alongside the
corresponding results.

\begin{table*}[t]
\centering
\small
\setlength{\tabcolsep}{3pt}
\caption{Training and validation performance
and latent-space organization on the
$\mathcal{D}^{(1)}$--$\mathcal{D}^{(3)}$ subsets.
Train Acc. and Val. Acc. denote training and validation accuracy,
respectively, reported as proportions.
Sil. Score and Align. Dist. denote the silhouette score and
simulation-to-experiment centroid alignment distance, both
calculated from validation pre-head embeddings.
Values are mean $\pm$ sample standard deviation across three
repeats. Higher accuracy and silhouette scores and lower
alignment distances indicate better performance.
Sim. and Exp. denote simulated and experimental data,
respectively.}
\label{tab:latent_metrics}
\begin{tabular*}{\textwidth}{@{\extracolsep{\fill}}lcccccc}
\hline
\textbf{Model}
& \multicolumn{2}{c}{\textbf{Train Acc.}}
& \multicolumn{2}{c}{\textbf{Val. Acc.}}
& \textbf{Sil. Score}
& \textbf{Align. Dist.} \\
\cline{2-3}
\cline{4-5}
& \textbf{Sim.}
& \textbf{Exp.}
& \textbf{Sim.}
& \textbf{Exp.}
& \textbf{Val.}
& \textbf{Val.} \\
\hline
CNN
& $0.9439 \pm 0.0483$
& $0.9423 \pm 0.0447$
& $0.9431 \pm 0.0398$
& $0.9414 \pm 0.0480$
& $0.215 \pm 0.059$
& $12.696 \pm 0.614$ \\
CNN--SNGP
& $0.9464 \pm 0.0443$
& $0.9454 \pm 0.0473$
& $0.9409 \pm 0.0456$
& $0.9474 \pm 0.0456$
& $0.533 \pm 0.007$
& $6.838 \pm 0.641$ \\
CNN--SNGP--adapt
& $0.9472 \pm 0.0455$
& $0.9569 \pm 0.0385$
& $0.9518 \pm 0.0341$
& $0.9569 \pm 0.0417$
& $0.664 \pm 0.078$
& $3.113 \pm 1.203$ \\
\hline
\end{tabular*}
\end{table*}

\begin{figure*}[!t]
    \centering
    \includegraphics[width=\textwidth]{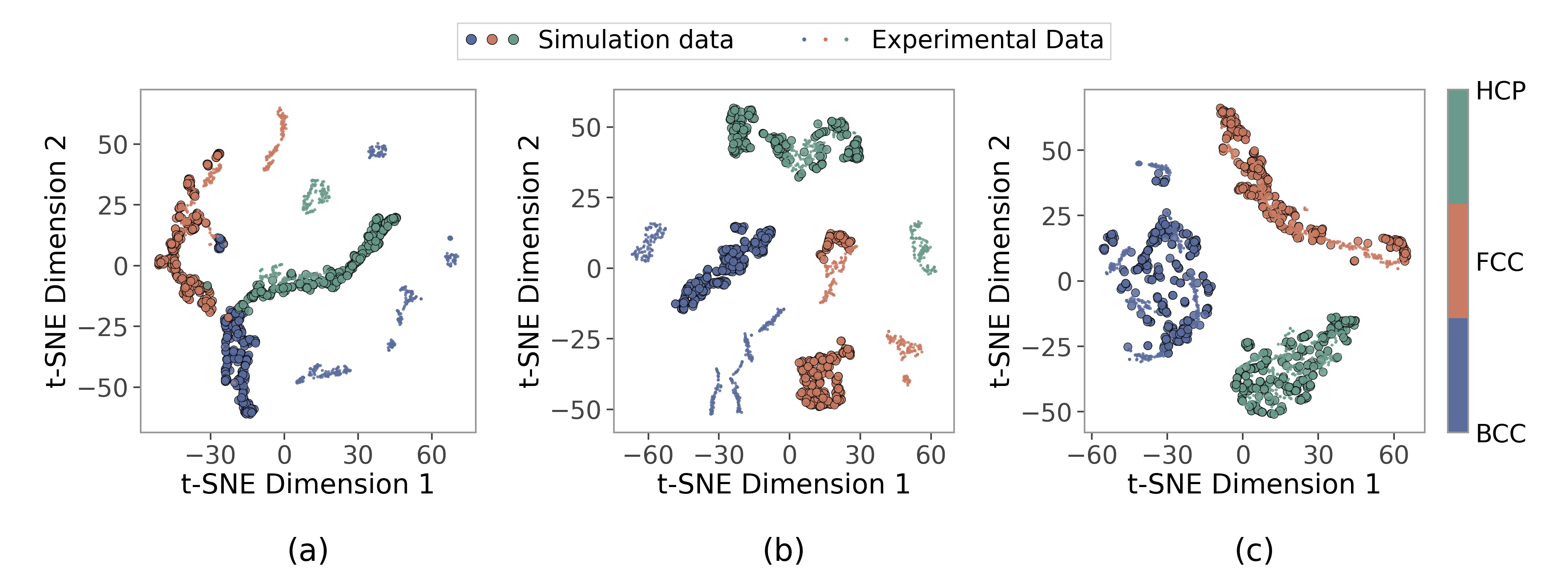}
    \caption{\textbf{t-SNE visualization of latent spaces learned by different models.}
    (a) A conventional CNN produces a weakly organized embedding with class overlap,
    (b) CNN--SNGP improves class separation but retains domain-specific separation between simulated and experimental samples,
    and (c) CNN--SNGP-adapt yields compact, class-consistent clusters with improved simulation-to-experiment alignment.}
    \label{fig:CNN-SNGP-adapt}
\end{figure*}

\subsection{Implementation and Training Details}

All experiments were run on a workstation with Ubuntu 20.04, a 24-core
AMD EPYC 7413 processor, 64 GB of RAM, and an NVIDIA RTX A6000 GPU.
We implemented the framework in Python 3.9 using
PyTorch~\cite{paszke2019pytorch}. We trained the fixed and continual-learning models with Adam using
mini-batches of 32 scans per domain and up to 800 epochs per fitting
attempt. Fresh fits used an initial learning rate of $5 \times 10^{-4}$.
Subsequent updates initialized
from the preceding stage's selected checkpoint used an initial learning
rate of $2 \times 10^{-5}$. The learning rate was multiplied by 0.997
after each epoch.

For the CNN--SNGP evaluations, spectral normalization is applied to \texttt{fc1} and \texttt{fc2}, with a nominal unit spectral norm estimated using one power iteration per forward call. This normalization is approximate rather than a certified bound. The GP output layer uses 64 random Fourier features (RFFs) to approximate the RBF kernel $k(h,h')=\exp(-\gamma\|h-h'\|^2)$, with $\gamma=1/64$, corresponding to a length scale of $\ell=\sqrt{32}$. RFF initialization uses the global random seed: 42 for the fixed evaluation and 42, 43, and 44 for the grouped repeats, without a separate RFF random-number generator. The ridge penalty is 1.0, and covariance momentum is set to \texttt{None}, enabling additive accumulation of precision statistics without an exponential moving average. Source and target batches each contribute one update, giving two updates per paired optimizer step. These statistics are reset each epoch and rebuilt from training inputs before evaluation, with the selected model weights held fixed.

Checkpoint selection and early stopping were based on validation
data. Early stopping used a patience of
30 epochs. For the continual-learning comparison, the validation
criterion averages cross-entropy across introduced blocks within
each domain and then sums the domain-specific means. The minimum
unsmoothed criterion selects the checkpoint, whereas an exponential
moving average with coefficient 0.2 controls stopping. Test data are excluded from model
selection, stopping decisions, and hyperparameter selection.

Labeled simulated samples and the revealed experimental training
anchors provide supervised classification information. The true
labels of the remaining experimental training samples are withheld;
these samples contribute through the applicable unlabeled-data
objectives. Experimental validation labels provide separate
model-selection supervision and are not included in the
training-anchor budget. Compared methods receive matched partitions,
anchor masks, and validation information, while held-out test
labels are used only for performance assessment.

\section{Results}\label{sec_result}

\subsection{Prediction Accuracy and Latent Space Distribution}

We first evaluate whether the proposed framework can infer crystal
structure from 2D XRD patterns and whether domain adaptation improves
the organization of the learned feature space. This comparison uses
a fixed, nonsequential training protocol. Three models are compared:
(i) a CNN classifier, (ii) CNN--SNGP, and (iii) a domain-adapted
CNN--SNGP model (CNN--SNGP--adapt). Each method is evaluated using
eight material-held-out folds across three repeats, yielding
24 model fits per method and 72 fits in total. We analyze the
$\mathcal{D}^{(1)}$--$\mathcal{D}^{(3)}$ subsets of these evaluations,
retaining their original training and validation assignments.
Reported values are the mean and sample standard deviation across
the three repeat-level estimates.

All three models use the same partitions, experimental training
anchors, and validation information. Labeled simulated XRD images
provide source-domain supervision, while experimental training
anchors provide partial target-domain supervision at a requested
labeling fraction of 10\%.
This fraction was selected based on the anchor-budget sensitivity
analysis reported in Appendix~\ref{app:anchor_budget_sensitivity},
where 10\% was the smallest tested budget attaining the
experimental validation-accuracy plateau.

The models share the same CNN encoder. CNN uses a linear
classification head, whereas CNN--SNGP uses a random-feature
mapping and GP output head. CNN--SNGP--adapt additionally
incorporates domain alignment.
The comparison therefore evaluates the complete training pipelines.

The three models are first evaluated in terms of classification
accuracy on simulated and experimental data. As shown in
Table~\ref{tab:latent_metrics}, all models achieve approximately
94--96\% accuracy on the training and validation subsets.
CNN--SNGP--adapt achieves the highest mean experimental validation
accuracy, $95.69 \pm 4.17\%$, compared with
$94.74 \pm 4.56\%$ for CNN--SNGP and
$94.14 \pm 4.80\%$ for CNN. It also achieves the highest mean
simulated validation accuracy. These development-set results
describe predictive performance on materials represented during
training; generalization to held-out materials is examined
in the following subsection.

Classification accuracy alone does not indicate whether the learned
feature space is well organized or whether simulated and experimental
representations are effectively aligned. We therefore further examine
the learned representations using the t-SNE visualizations in
Figure~\ref{fig:CNN-SNGP-adapt}. We randomly select one of the
24 material-grouped runs and use the same run index, corresponding
to the same material holdout and repeat, for all three models.
For visualization purposes only, the raw 256-dimensional pre-head
features are projected into two dimensions. Each panel is fitted
independently, so absolute positions and inter-cluster distances
are not directly comparable across panels. These plots provide
qualitative illustrations for the selected run rather than
quantitative measures of domain alignment or summaries of
all evaluations.

To quantitatively assess the learned latent space, we evaluate
the silhouette score and simulation-to-experiment centroid
alignment distance, as defined in Section~\ref{sec:metrics}.
Both metrics use the common 256-dimensional pre-head
representations. The silhouette score measures class separation,
whereas the alignment distance measures the average separation
between simulated and experimental centroids within each crystal
class. The values in Table~\ref{tab:latent_metrics} are calculated
from the $\mathcal{D}^{(1)}$--$\mathcal{D}^{(3)}$ validation
embeddings.

As shown in Table~\ref{tab:latent_metrics}, both SNGP-based models
achieve higher validation silhouette scores and lower alignment
distances than CNN. CNN--SNGP--adapt achieves the highest mean
silhouette score, $0.664 \pm 0.078$, compared with
$0.533 \pm 0.007$ for CNN--SNGP and
$0.215 \pm 0.059$ for CNN. Its mean alignment distance is also
the lowest, decreasing to $3.113 \pm 1.203$ from
$6.838 \pm 0.641$ for CNN--SNGP and
$12.696 \pm 0.614$ for CNN. These results indicate stronger
class separation and closer class-conditional simulated--experimental
centroids in the adapted model's validation representations.

\begin{table*}[t]
\centering
\small
\setlength{\tabcolsep}{3pt}
\caption{Classification and predictive-probability quality on
material-held-out data. Within each repeat, predictions are pooled across the eight
material holdouts before calculating metrics. Values are
mean $\pm$ sample standard deviation across the three
repeat-level estimates. Acc., Bal. Acc., and F1 denote
accuracy, balanced accuracy, and macro-F1. NLL denotes negative
log-likelihood, Brier denotes the multiclass Brier score, and
ECE denotes expected calibration error. Higher classification
metrics and lower NLL, Brier, and ECE indicate better performance.
CNN denotes convolutional neural network, SNGP denotes
spectral-normalized neural Gaussian process, and adapt denotes
the domain-adapted pipeline.}
\label{tab:heldout_accuracy}

\begin{tabular*}{\textwidth}{@{\extracolsep{\fill}}lcccccc}
\multicolumn{7}{l}{
\textbf{(A) Material-held-out classification}} \\
\hline
\textbf{Model}
& \multicolumn{3}{c}{\textbf{Simulated}}
& \multicolumn{3}{c}{\textbf{Experimental}} \\
\cline{2-4}
\cline{5-7}
& \textbf{Acc. (\%)}
& \textbf{Bal. Acc. (\%)}
& \textbf{F1}
& \textbf{Acc. (\%)}
& \textbf{Bal. Acc. (\%)}
& \textbf{F1} \\
\hline
CNN
& $59.1 \pm 11.4$
& $56.4 \pm 11.3$
& $0.541 \pm 0.135$
& $29.1 \pm 10.2$
& $28.8 \pm 8.7$
& $0.290 \pm 0.069$ \\
CNN--SNGP
& $64.9 \pm 9.2$
& $63.0 \pm 9.4$
& $0.621 \pm 0.105$
& $44.3 \pm 4.9$
& $42.6 \pm 6.5$
& $0.421 \pm 0.059$ \\
CNN--SNGP--adapt
& $61.6 \pm 11.9$
& $58.4 \pm 13.3$
& $0.562 \pm 0.170$
& $50.3 \pm 6.9$
& $48.2 \pm 6.6$
& $0.499 \pm 0.067$ \\
\hline
\end{tabular*}

\medskip

\begin{tabular*}{\textwidth}{@{\extracolsep{\fill}}lccc}
\multicolumn{4}{l}{
\textbf{(B) Experimental predictive-probability quality}} \\
\hline
\textbf{Model}
& \textbf{NLL}
& \textbf{Brier}
& \textbf{ECE (\%)} \\
\hline
CNN
& $92.55 \pm 31.40$
& $1.302 \pm 0.243$
& $65.2 \pm 12.3$ \\
CNN--SNGP
& $3.02 \pm 0.87$
& $0.972 \pm 0.161$
& $51.0 \pm 12.4$ \\
CNN--SNGP--adapt
& $2.51 \pm 0.41$
& $0.861 \pm 0.065$
& $41.5 \pm 8.5$ \\
\hline
\end{tabular*}
\end{table*}

\begin{table*}[t]
\centering
\small
\setlength{\tabcolsep}{3pt}
\caption{Detection performance for noise-only and silicon carbide
(SiC) inputs. Out-of-distribution (OOD) inputs are treated as
the positive class. AUROC and AUPR denote the areas under the
receiver operating characteristic and precision--recall curves.
FPR95 is the false-positive rate on known-class reference inputs
at an operating point attaining at least 95\% OOD detection.
OOD false acceptance is the fraction of OOD inputs accepted
at the evaluated rejection operating point.
Higher AUROC and AUPR and lower error rates indicate better
performance. Values are mean $\pm$ standard deviation across repeated runs.
CNN denotes convolutional neural network, SNGP denotes
spectral-normalized neural Gaussian process, and adapt denotes
the domain-adapted pipeline.}
\label{tab:ood_detection}
\begin{tabular*}{\textwidth}{@{\extracolsep{\fill}}llcccc}
\hline
\textbf{Input}
& \textbf{Model}
& \textbf{AUROC}
& \textbf{AUPR}
& \textbf{FPR95 (\%)}
& \shortstack{\textbf{OOD false}\\\textbf{acceptance (\%)}} \\
\hline
Noise
& CNN
& $0.721 \pm 0.098$
& $0.512 \pm 0.274$
& $37.320 \pm 4.221$
& $100.000 \pm 0.000$   \\
Noise
& CNN--SNGP
& $0.998 \pm 0.001$
& $0.989 \pm 0.006$
& $0.816 \pm 0.019$
& $0.000 \pm 0.000$\\
Noise
& CNN--SNGP--adapt
& $1.000 \pm 0.000$ 
& $1.000 \pm 0.000$ 
& $0.000 \pm 0.000$ 
& $0.000 \pm 0.000$  \\
\hline
SiC
& CNN
& $0.749 \pm 0.103$
& $0.541 \pm 0.202$
& $32.165 \pm 6.332$
& $100.000 \pm 0.000$  \\
SiC
& CNN--SNGP
& $0.997 \pm 0.001$
& $0.992 \pm 0.003$
& $0.712 \pm 0.114$
& $0.000 \pm 0.000$\\
SiC
& CNN--SNGP--adapt
& $1.000 \pm 0.000$ 
& $1.000 \pm 0.000$ 
& $0.000 \pm 0.000$ 
& $0.000 \pm 0.000$  \\
\hline
\end{tabular*}
\end{table*}

\subsection{Prediction Accuracy and Calibration on Unseen Materials}
\label{subsec:unseen_materials}

To assess transfer beyond the materials represented in training,
we perform repeated leave-one-material-out evaluation for Al, Cu,
Fe, Mo, Nb, Ni, Ti, and Zn. Each held-out material is excluded
from training and validation in both simulated and experimental
domains.
The held-out materials belong to the known BCC/FCC/HCP classes;
this protocol therefore evaluates transfer to unfamiliar material
identities within the existing label space.

Each model is evaluated across eight material holdouts and three
repeats, yielding 24 fits per model. Within each repeat, predictions
across the eight holdouts are pooled before calculating performance
metrics. We report the mean and sample standard deviation of the
three repeat-level estimates. CNN, CNN--SNGP, and CNN--SNGP-adapt
use matched partitions, experimental training anchors, and
validation information.

As shown in Table~\ref{tab:heldout_accuracy}, panel~A,
CNN--SNGP-adapt achieves the highest mean experimental accuracy
at $50.3 \pm 6.9\%$, followed by CNN--SNGP at
$44.3 \pm 4.9\%$ and CNN at $29.1 \pm 10.2\%$.
Experimental balanced accuracy and macro-F1 follow the same
ordering. Paired material-level analysis estimates accuracy gains
of 21.2 percentage points over CNN and 6.0 percentage points over
CNN--SNGP. The corresponding 95\% class-stratified material-bootstrap
confidence intervals are $[2.8,41.5]$ and $[0.8,13.1]$ percentage
points, respectively, conditional on the observed material groups
and fitted runs.

We also evaluate predictive-probability quality, since improved
accuracy does not necessarily imply reliable confidence.
CNN uses ordinary softmax probabilities, whereas the SNGP models
use analytical variance-adjusted mean-field probabilities.
As shown in Table~\ref{tab:heldout_accuracy}, panel~B, both
SNGP-based models achieve lower mean NLL, Brier score, and ECE
than CNN on experimental holdouts, with CNN--SNGP-adapt obtaining
the lowest values.

\subsection{Detection of Noise-Only and Out-of-Label-Space Inputs}
\label{subsec:OOD}

We separately evaluate whether predictive confidence can
distinguish valid known-class inputs from noise-only and
SiC inputs. Noise-only inputs lack useful diffraction
information, whereas SiC lies outside the trained
BCC/FCC/HCP structure-label set. Neither input type has a
valid target label within this classification task.
We therefore evaluate unfamiliar-input detection rather
than three-class classification accuracy.

Table~\ref{tab:ood_detection}
reports AUROC, AUPR, FPR95, and OOD false acceptance
as mean $\pm$ standard deviation across repeated runs
for the evaluated datasets. As shown in Table~\ref{tab:ood_detection}, both SNGP-based
models provide stronger discrimination between known-class
reference inputs and the evaluated OOD inputs than CNN.
CNN--SNGP achieves AUROC values of 0.998 for noise and
0.997 for SiC, with FPR95 below 1\% for both input types.
At the reported precision, CNN--SNGP--adapt achieves AUROC
and AUPR values of 1.000 and zero FPR95 for both input types.

At the evaluated rejection operating point, neither
SNGP-based model accepts any evaluated OOD inputs,
whereas CNN accepts all of them. This false-acceptance
result is distinct from FPR95: FPR95 measures rejection
of known-class inputs at an operating point attaining
at least 95\% OOD detection, while OOD false acceptance
measures acceptance of unfamiliar inputs at the reported
rejection threshold.

Overall, these results demonstrate that SNGP-based uncertainty estimation substantially improves the detection of unfamiliar XRD inputs compared with a conventional CNN. The adaptation-enhanced model further improves discrimination, achieving near-perfect OOD detection on the evaluated noise-only and SiC datasets. These findings suggest that integrating uncertainty estimation with simulation-to-experiment adaptation can support more reliable XRD interpretation by identifying inputs that should not receive confident predictions within the trained label space.

\begin{figure*}[p]
\centering
\includegraphics[width=0.95\textwidth]{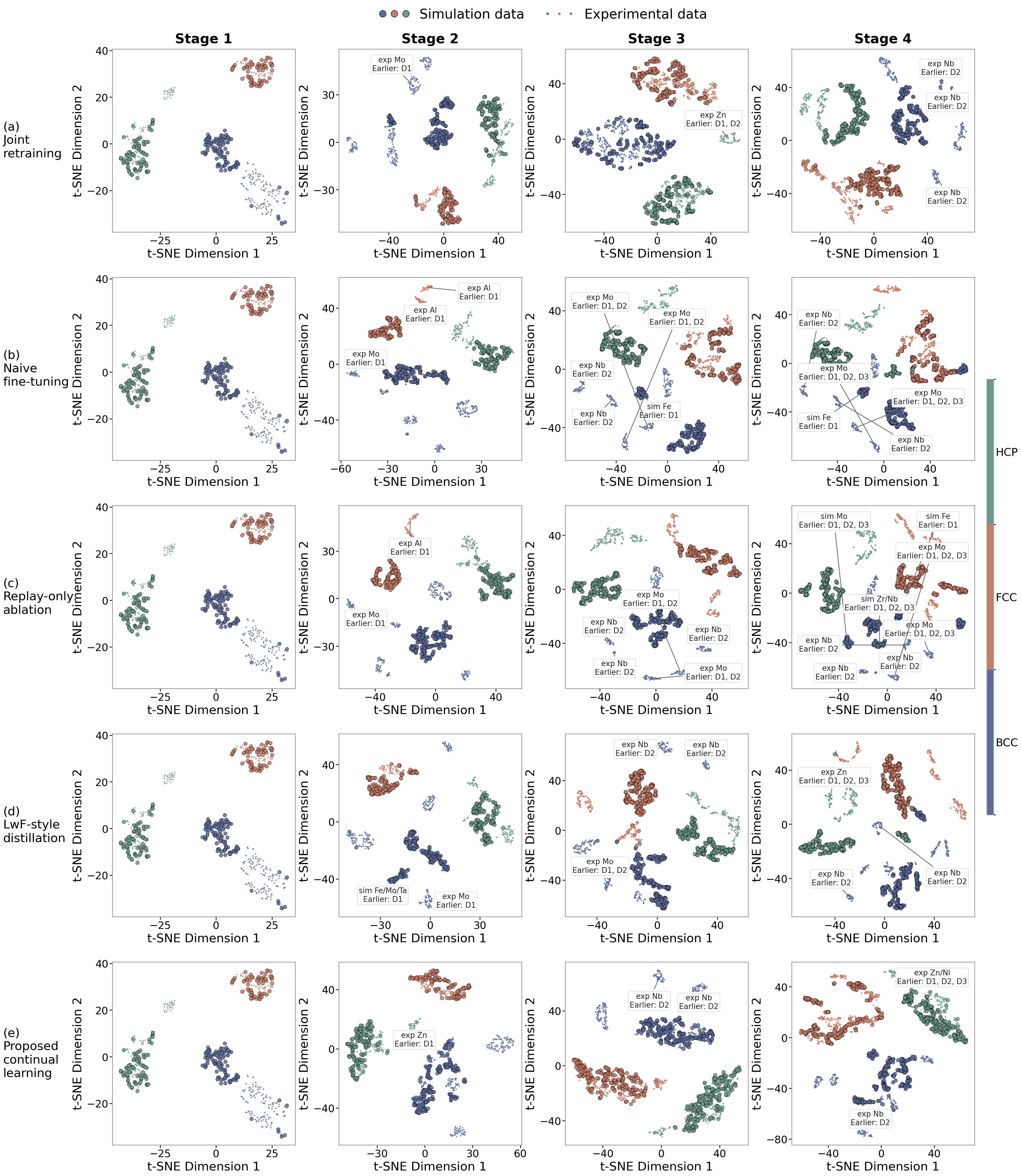}
\caption{\textbf{Pre-head latent-space evolution during sequential learning.}
Rows show (a) joint retraining, (b) na\"ive fine-tuning,
(c) replay-only updating, (d) learning-without-forgetting
(LwF)-style distillation, and (e) the proposed continual-learning
strategy; columns show stages 1--4.
At each stage, all methods are visualized using the same ordered
set of accumulated training, validation, and test scans.
Blue, orange, and green indicate body-centered cubic (BCC),
face-centered cubic (FCC), and hexagonal close-packed (HCP)
structures, respectively.
Larger black-edged circles denote simulated scans, and small dots
denote experimental scans.
Annotations identify selected material subsets and the earlier
blocks in which they appeared.
Each t-SNE panel is fitted independently; absolute positions and
distances should not be compared across panels.}
\label{fig:no_continual}
\end{figure*}

\begin{figure*}[t]
\centering
\includegraphics[width=\textwidth]{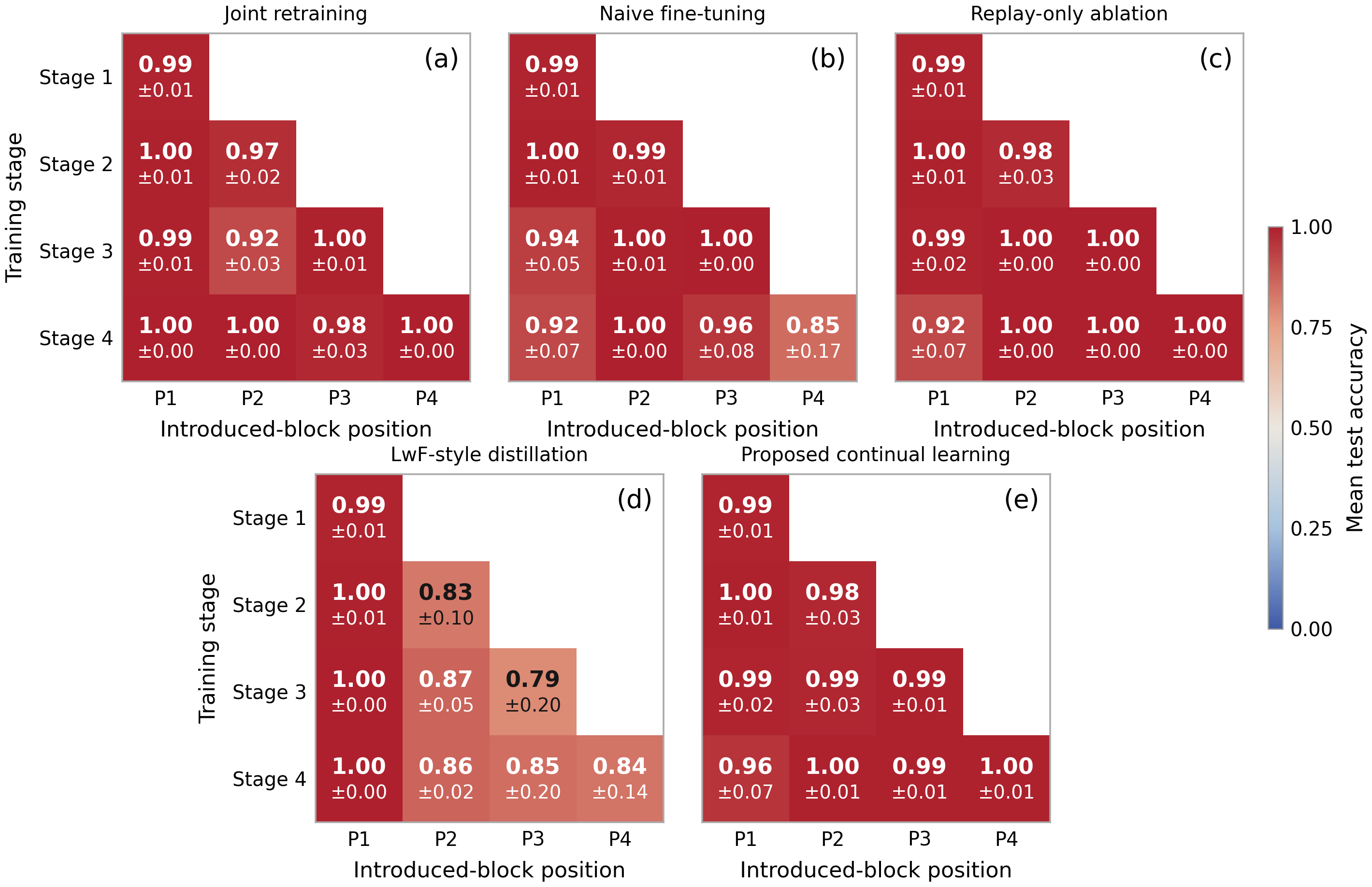}
\caption{\textbf{Stage-wise test accuracy under five sequential
training strategies.}
Panels show (a) joint retraining, (b) na\"ive fine-tuning,
(c) replay-only updating, (d) learning-without-forgetting
(LwF)-style distillation, and (e) the proposed continual-learning
strategy.
Rows indicate model-update stages, and columns indicate
introduced-block positions (P1--P4) within each evaluated sequence.
Block identities at these positions depend on the introduction order.
Each cell displays the mean test accuracy, with the standard
deviation beneath it.
Future-block cells are blank.
Accuracies are proportions rounded to two decimals, and all panels
use a common 0--1 color scale.}
\label{fig:accuracy_matrix}
\end{figure*}

\subsection{Continual Learning for Autonomous XRD Analysis}
\label{subsec:continual_learning_current_results}

\begin{table*}[t]
\centering
\small
\setlength{\tabcolsep}{3pt}
\caption{
Final-stage classification performance, latent-space diagnostics,
and cumulative training time for five primary strategies.
Results average orders within each seed and report mean
$\pm$ sample standard deviation across three seeds.
Accuracy pools test scans from all introduced blocks.
Sim. and Exp. denote simulated and experimental data.
$S_{\mathrm{raw}}$ denotes the raw validation silhouette score;
$D_{\mathrm{align,std}}$ denotes the standardized source--target
class-centroid gap. Arrows indicate the preferred direction.
CL denotes continual learning.
Training time includes the complete sequence, validation, GP precision reconstruction, and retries, excluding
test exports and separate plotting.}
\label{tab:continual_metrics}

\begin{tabular*}{\textwidth}{@{\extracolsep{\fill}}lccccc}
\hline
\textbf{Strategy}
& \shortstack{\textbf{Sim. Acc.}\\\textbf{(\%)}}
& \shortstack{\textbf{Exp. Acc.}\\\textbf{(\%)}}
& $S_{\mathrm{raw}}\uparrow$
& $D_{\mathrm{align,std}}\downarrow$
& \shortstack{\textbf{Training time}\\\textbf{(h)}} \\
\hline
Joint retraining
& $99.88 \pm 0.22$
& $99.44 \pm 0.97$
& $0.904 \pm 0.020$
& $1.156 \pm 0.078$
& $8.49 \pm 1.99$ \\
Na\"ive fine-tuning
& $95.88 \pm 3.20$
& $94.23 \pm 3.05$
& $0.408 \pm 0.116$
& $6.068 \pm 1.013$
& $2.32 \pm 0.27$ \\
Replay-only
& $99.75 \pm 0.43$
& $97.47 \pm 2.19$
& $0.616 \pm 0.190$
& $4.165 \pm 1.754$
& $4.69 \pm 0.37$ \\
LwF-style distillation
& $93.51 \pm 2.49$
& $89.31 \pm 0.24$
& $0.353 \pm 0.049$
& $5.787 \pm 0.834$
& $1.63 \pm 0.24$ \\
Proposed CL
& $99.63 \pm 0.37$
& $98.31 \pm 1.93$
& $0.761 \pm 0.066$
& $2.788 \pm 0.594$
& $4.36 \pm 0.65$ \\
\hline
\end{tabular*}
\end{table*}

\begin{table*}[t]
\centering
\small
\setlength{\tabcolsep}{3pt}
\caption{
Final experimental-domain retention and transfer metrics,
using the aggregation in Table~\ref{tab:continual_metrics}.
Average accuracy (AA) weights blocks equally.
Forgetting, backward transfer (BWT), and forward transfer (FWT)
are in percentage points (pp).
Old and New denote task-average accuracy on earlier blocks
and accuracy on the latest introduced block, respectively.
Negative signed forgetting indicates improvement beyond the
best earlier post-introduction accuracy. BWT references accuracy
at introduction; FWT references the untrained model.
CL denotes continual learning.}
\label{tab:continual_transfer_calibration}

\begin{tabular*}{\textwidth}{@{\extracolsep{\fill}}lcccccc}
\hline
\textbf{Strategy}
& \shortstack{\textbf{AA}\\\textbf{(\%)}}
& \shortstack{\textbf{Forgetting}\\\textbf{(pp)}}
& \shortstack{\textbf{BWT}\\\textbf{(pp)}}
& \shortstack{\textbf{FWT}\\\textbf{(pp)}}
& \shortstack{\textbf{Old}\\\textbf{(\%)}}
& \shortstack{\textbf{New}\\\textbf{(\%)}} \\
\hline
Joint retraining
& $99.56 \pm 0.77$
& $-0.15 \pm 1.68$
& $0.89 \pm 0.77$
& $29.12 \pm 23.63$
& $99.41 \pm 1.03$
& $100.00 \pm 0.00$ \\
Na\"ive fine-tuning
& $93.11 \pm 3.40$
& $4.00 \pm 4.00$
& $-3.41 \pm 3.83$
& $26.13 \pm 20.95$
& $95.85 \pm 4.22$
& $84.89 \pm 16.88$ \\
Replay-only
& $98.00 \pm 1.73$
& $2.52 \pm 2.19$
& $-1.63 \pm 3.37$
& $20.97 \pm 19.57$
& $97.33 \pm 2.31$
& $100.00 \pm 0.00$ \\
LwF-style distillation
& $88.63 \pm 2.65$
& $-0.63 \pm 1.67$
& $3.39 \pm 3.34$
& $14.55 \pm 23.04$
& $90.33 \pm 6.20$
& $83.56 \pm 14.26$ \\
Proposed CL
& $98.63 \pm 1.52$
& $0.84 \pm 2.41$
& $-0.55 \pm 2.78$
& $22.38 \pm 18.72$
& $98.32 \pm 2.08$
& $99.56 \pm 0.77$ \\
\hline
\end{tabular*}
\end{table*}

\begin{table}[t]
\centering
\small
\setlength{\tabcolsep}{3pt}
\caption{
Final experimental predictive-probability quality for five
primary strategies using the same CNN--SNGP--adapt backbone.
Results use the aggregation in Table~\ref{tab:continual_metrics}.
NLL denotes negative log-likelihood, ECE denotes expected
calibration error, and CL denotes continual learning.
ECE uses 15 equal-width confidence bins; NLL uses natural
logarithms. Lower values indicate better performance.
Probabilities use the analytical mean-field calculation
without post-hoc temperature scaling.}
\label{tab:continual_probability}

\begin{tabular*}{\columnwidth}{@{\extracolsep{\fill}}lcc}
\hline
\textbf{Strategy}
& \textbf{NLL}
& \textbf{ECE (\%)} \\
\hline
Joint retraining
& $0.017 \pm 0.025$
& $0.88 \pm 1.12$ \\
Na\"ive fine-tuning
& $0.228 \pm 0.103$
& $12.03 \pm 3.97$ \\
Replay-only
& $0.100 \pm 0.045$
& $6.82 \pm 2.15$ \\
LwF-style distillation
& $0.391 \pm 0.035$
& $19.38 \pm 5.37$ \\
Proposed CL
& $0.088 \pm 0.024$
& $6.58 \pm 1.67$ \\
\hline
\end{tabular*}
\end{table}

To evaluate the ability of the proposed framework to
incorporate new XRD measurements while retaining previously
learned knowledge, we compare five sequential training
strategies: joint retraining, na\"ive fine-tuning,
replay-only updating, learning-without-forgetting (LwF)-style
distillation, and the proposed continual-learning (CL)
strategy. The evaluation follows a fixed-label,
domain-incremental setting in which the BCC/FCC/HCP
classification space remains unchanged as new experimental
data become available. Experimental blocks $D_1$--$D_3$
share Al, Mo, Ti, and Zn, with Fe, Nb, and Ni introduced
in successive blocks, while $D_4$ contains FCC Cu only.
All simulated blocks contain all three crystal classes.
Although recurring materials have distinct image identities,
scans from the same specimen may remain correlated.
Accordingly, this evaluation focuses on sequential
adaptation within known crystal classes rather than
class-incremental learning or transfer across instruments
and acquisition campaigns.

For a controlled comparison, all strategies use the same
CNN--SNGP--adapt architecture, preprocessing, experimental
supervision, and validation information. Joint retraining uses all accumulated
training data, whereas na\"ive fine-tuning updates the
model using only the current block. Replay-only updating
retains the proposed memory-selection policy without
distillation, while LwF-style distillation uses current
inputs without replay memory. The proposed CL strategy
combines bounded replay with logit and feature distillation to
balance adaptation and knowledge retention. The replay
memory contains 128 images, equally divided between
simulated and experimental data, with memory-capacity
sensitivity examined in
Appendix~\ref{app:replay_memory_sensitivity}.
Hidden experimental training labels are excluded from
optimization and replay, and validation data are
maintained separately from the replay memory.

To assess robustness to initialization and data order,
each strategy is evaluated using three random seeds
(42--44) and three block sequences:
$(D_1,D_2,D_3,D_4)$, $(D_2,D_3,D_4,D_1)$, and
$(D_4,D_3,D_2,D_1)$. This produces 45 complete
four-stage training sequences across the five strategies.
Sample assignments and experimental anchors are matched
within each seed--order combination. Results are first
averaged across block orders within each seed and then
reported as mean $\pm$ sample standard deviation
across the three seeds.

Table~\ref{tab:continual_metrics} compares final-stage
classification accuracy, latent-space characteristics,
and cumulative training time. Joint retraining achieves
the highest accuracy but requires the greatest
computational effort. The proposed CL strategy approaches
this performance with substantially lower training time,
while outperforming na\"ive fine-tuning and LwF-style
distillation in both simulated and experimental accuracy.
Compared with replay-only updating, the proposed strategy
achieves slightly higher experimental accuracy and
slightly lower simulated accuracy, indicating that
combining replay and distillation offers a favorable performance improvement.


The latent-space diagnostics in
Table~\ref{tab:continual_metrics} provide further insight
into the learned representations. Compared with
fine-tuning and the two component controls, the proposed
CL strategy produces better class separation and a
smaller source--target class-centroid gap, although joint
retraining remains the strongest overall. These trends
are illustrated in Figure~\ref{fig:no_continual},
which visualizes the feature distributions for the
reference block order and seed 42. Selected detached
groups are annotated to identify their material
composition and association with earlier training
blocks. While these visualizations help interpret
changes in the learned representations, detached
groups alone do not establish forgetting, and distances
between independently fitted t-SNE projections
should not be compared directly.

To examine knowledge retention throughout sequential
training, Figure~\ref{fig:accuracy_matrix} presents
stage-wise test accuracy on previously introduced and
newly encountered blocks. The corresponding quantitative
results are summarized in
Table~\ref{tab:continual_transfer_calibration}.
Average accuracy (AA) weights all blocks equally,
whereas the accuracy in
Table~\ref{tab:continual_metrics} is pooled across
individual test scans. Forgetting measures the change
relative to the best earlier post-introduction
performance, backward transfer (BWT) measures changes
relative to performance at block introduction, and
forward transfer (FWT) compares pre-introduction
performance against the corresponding untrained model.
Future-block evaluations are used only for transfer
analysis and do not influence model fitting or
checkpoint selection.

The proposed CL strategy achieves higher accuracy on
both earlier and newly introduced blocks than na\"ive
fine-tuning, while exhibiting substantially less
forgetting. Relative to replay-only updating, the
proposed strategy improves retention of earlier blocks,
with a small reduction in accuracy on the newest block.
Although LwF-style distillation exhibits limited
forgetting and positive BWT, its lower overall accuracy
indicates that favorable retention metrics do not
necessarily imply effective sequential learning.
Forward-transfer estimates also exhibit considerable
variation, preventing a consistent advantage from
being established for the proposed strategy.


Beyond classification accuracy and retention,
maintaining reliable predictive probabilities is
important for uncertainty-aware XRD interpretation.
We therefore evaluate negative log-likelihood (NLL),
Brier score, and expected calibration error (ECE). The Gaussian-process
precision matrix is reconstructed using only permitted
training and replay inputs, without incorporating
validation or test data. As shown in Table~\ref{tab:continual_probability},
joint retraining achieves the lowest mean NLL and ECE,
whereas LwF-style distillation performs the worst.
The proposed CL strategy substantially improves both
metrics compared with na\"ive fine-tuning, indicating
that the benefits of replay and distillation extend
beyond classification accuracy to predictive-probability
quality. It also achieves slightly lower mean NLL and
ECE than replay-only updating.

Overall, the results demonstrate that the proposed
CL strategy provides an effective compromise between
predictive performance, knowledge retention, and
computational efficiency. Although joint retraining
remains the strongest reference, the proposed strategy
achieves comparable classification accuracy at lower
training cost and improves retention and
predictive-probability quality relative to na\"ive
fine-tuning. These findings support the use of bounded
replay and distillation for incrementally updating
XRD classification models as new measurements
become available.

\section{Discussion}
\label{sec_discussion}

\subsection{Contribution and Relevance to Autonomous Experimentation}
\label{subsec:novelty}

The primary contribution of this work is the integration and evaluation of established learning components for uncertainty-aware, sequential interpretation of image-based characterization data. Prior work has demonstrated that simulated diffraction data can support experimental XRD analysis and that uncertainty estimates can inform autonomous characterization \cite{szymanski2023adaptively,chen2026uncertainty}. Building on these capabilities, this study combines CNN--SNGP uncertainty estimation, MMD-based domain adaptation, and replay with knowledge distillation. The 2D XRD application demonstrates how these components can support cross-domain prediction and knowledge retention as additional data are introduced.

The experiments are conducted offline using predefined data blocks, with model updates performed at block boundaries. The model does not select subsequent measurements, trigger rescanning, or modify instrument settings. This evaluation therefore concerns characterization-model updating rather than end-to-end autonomous experimentation. Its relevance to autonomous laboratories lies in maintaining predictive capabilities as additional measurements become available and data distributions evolve \cite{stach2021autonomous}.

Future integration into an AES could support materials discovery
by coupling XRD interpretation with machine-learning models
linking elemental composition to material properties. Structural
predictions from XRD could complement composition and measured
property data, while predictive uncertainty could guide the
selection of new compositions for synthesis and evaluation. 
For example, the AES could balance exploring uncertain regions
of composition space with identifying compositions likely to
improve a target property. Such integration would require
evaluating how characterization uncertainty affects experimental
selection and whether sequential model updates improve discovery
efficiency. These capabilities remain future extensions.

\subsection{Framework Scope}
\label{subsec:framework_scope}

This study establishes an offline proof-of-concept for processing 2D diffraction images while supporting uncertainty estimation, simulation-to-experiment adaptation, and continual model updating. SNGP combines neural-network feature extraction with predictive uncertainty estimation, MMD-based alignment addresses differences between simulated and experimental feature distributions, and replay with knowledge distillation supports updating within the fixed BCC/FCC/HCP label space.

The goal is to establish a modular framework rather than identify a single best-performing model. To provide context for the choice of SNGP, we conducted a simple comparison with a Deep Kernel Learning (DKL) \cite{wilson2016deep} using the same CNN feature extractor and training and test datasets, as reported in Appendix~\ref{app:supp_cnn_dkl}. This comparison examines a closely related uncertainty-estimation approach within the evaluated setting and does not establish the general superiority of either method.

The anchor-budget analysis in Appendix~C examines how the amount
of labeled experimental training data affects performance within
the proposed framework. Broader comparisons with source-only
training, target-only training, and simple target fine-tuning,
together with systematic ablations of alignment
objectives, would further clarify the contributions of simulation
data, experimental supervision, and individual adaptation components.

The current encoder serves as a working implementation for evaluating the integrated learning workflow. Its substantial parameter count carries storage and memory costs that should be considered in application-specific deployment. The reported training-time comparison evaluates continual updating relative to joint retraining using the same backbone; it does not establish that this backbone is the most computationally efficient architecture for the task. Pooling-based or lightweight encoders could be incorporated to meet different computational requirements, followed by retraining and validation.

The BCC/FCC/HCP classification task provides a controlled setting for demonstrating these integrated capabilities. Azimuthally integrated profiles and peak-based features can also support classification of these structures. Our motivation for using 2D inputs is to demonstrate direct image-based interpretation while preserving flexibility for broader characterization objectives. A two-dimensional representation retains the spatial organization of diffraction features, including azimuthal intensity variations and spot distributions, which may be relevant to future tasks involving texture or other microstructural characteristics. The present application therefore serves as a starting point for extending the framework to more varied image-based scientific data.

Beyond DKL, candidate uncertainty-estimation approaches include Bayesian neural networks \cite{kononenko1989bayesian}, Monte Carlo dropout \cite{seoh2020qualitative}, deep ensembles \cite{ganaie2022ensemble}, and Dirichlet-based models \cite{kopetzki2021evaluating}. Possible domain-adaptation approaches include adversarial methods \cite{sicilia2023domain,tzeng2017adversarial,long2018conditional}, feature-distribution alignment \cite{sun2016deep,long2015learning}, optimal-transport-based adaptation \cite{damodaran2018deepjdot}, and self-training or pseudo-labeling \cite{saito2017asymmetric,liu2021cycle,mei2020instance}. Their incorporation would require appropriate changes to training objectives and evaluation of predictive performance, uncertainty calibration, knowledge retention, and computational cost. Extension to other characterization modalities would similarly require suitable input representations, encoders, and task-specific validation.

\subsection{Limitations and Deployment Considerations}
\label{subsec:limitations}

Several limitations remain. First, this study is a proof-of-concept
covering a limited range of materials and experimental conditions.
Both the simulated and experimental datasets focus on single-phase
materials with BCC, FCC, or HCP structures. This controlled benchmark
supports evaluation of simulation-to-experiment transfer, predictive
uncertainty, and knowledge retention.
The model predicts a BCC, FCC, or HCP structure label from a single-phase XRD pattern, including for material identities excluded from training, as evaluated in Section \ref{subsec:unseen_materials}. The output labels do not distinguish compounds or specific phases sharing a structure type, and the model does not perform quantitative phase analysis, lattice-parameter refinement, or defect and strain characterization. Broader application would require validation on
lower-symmetry structures, multiphase systems, amorphous materials,
and patterns affected by texture or severe peak overlap.

The complexity required for useful AES operation
depends on the experimental objective rather than a universal minimum
number of classes. A characterization model must resolve the
structures, phases, or properties relevant to the intended experimental
decision. Expanding single-phase classification would require
representative data for additional classes and a corresponding
classification output. Phase identification would also require
distinguishing relevant compounds, including those sharing the same
symmetry. Multiphase interpretation would require representative
mixture data and simultaneous phase predictions, while quantitative
analysis would require outputs such as phase fractions or lattice
parameters. These extensions would require renewed evaluation of
domain alignment, predictive uncertainty, and knowledge retention
during sequential updates. The modular framework provides a basis
for investigating these tasks, although their performance and
computational demands remain to be evaluated. Comparisons with
simpler classifiers and conventional diffraction-analysis methods,
considering accuracy, memory use, and training and inference times,
would help determine when the integrated components provide sufficient
benefit to justify their additional cost.

Second, the continual-learning component is evaluated under controlled sequential updates within a fixed BCC/FCC/HCP label space. Expanding the output space beyond BCC, FCC, and HCP would require an explicit class-incremental learning strategy and remains untested. Even within a fixed label space, larger distribution shifts and longer update sequences may make it more difficult to balance adaptation against retention of earlier predictive capabilities. Restricting replay can limit the historical data processed during each update, but a fixed replay budget may become less representative as material and measurement diversity increase. Future evaluation should therefore examine accuracy, uncertainty calibration, forgetting, memory requirements, and update time as the number of classes, data distributions, and sequential updates grows. Such evaluation is necessary to establish whether the observed benefits persist under sustained autonomous operation.

The loss weights and distillation temperature were selected through preliminary training runs to provide a working configuration for evaluating the integrated framework. The weights balance classification, domain alignment, and knowledge retention through replay and distillation. Specifically, the distillation weight controls the emphasis on matching the preceding model's predictions, while the temperature adjusts the softness of the distillation probabilities. These choices can influence the balance between learning from new data and retaining prior knowledge. As this study focuses on the framework's integrated behavior, a comprehensive assessment of parameter sensitivity remains a direction for future work.

Finally, deployment in an autonomous laboratory requires validation beyond offline predictive performance. Inference and model updates must fit within the experimental timescale, and incoming data require quality checks for missing metadata, corrupted images, and changes in instrument settings. Predictive uncertainty must also be connected to validated response rules. Depending on the experimental objective, an uncertain prediction could trigger a repeated scan, a change in exposure time, complementary characterization, or human review. These actions are prospective integration options and have not been evaluated in a closed-loop experiment in this study. Deployment should therefore combine automated data-quality checks, monitoring of predictive performance, and predefined uncertainty-based actions, followed by experimental validation of their effects on characterization reliability and workflow efficiency.

\section{Conclusions}\label{sec_conclusion}

\renewcommand{\thefigure}{A\arabic{figure}}
\setcounter{figure}{0}

\renewcommand{\thetable}{A\arabic{table}}
\setcounter{table}{0}

\begin{figure*}[t]
    \centering
    \includegraphics[width=0.8\textwidth]
    {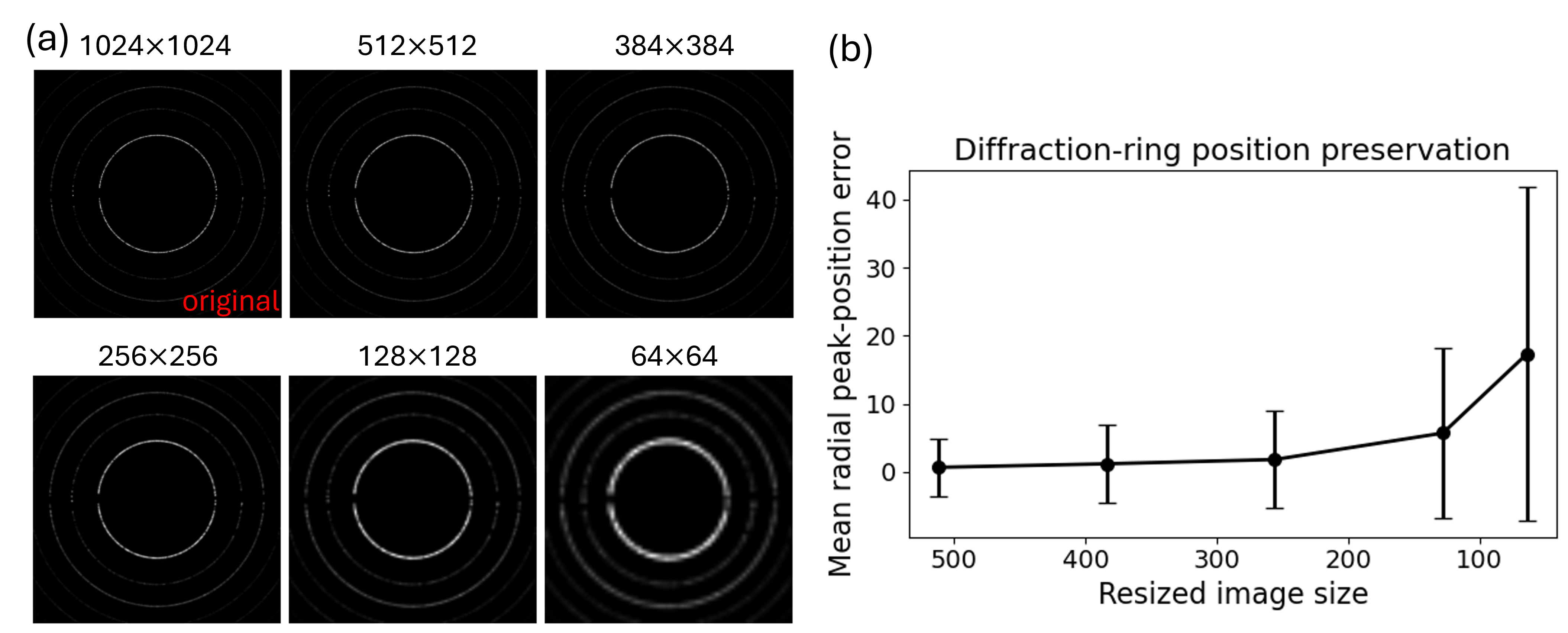}
    \caption{\textbf{Resolution-convergence analysis for XRD image resizing.}
    (a) Representative original and resized-restored XRD images.
    (b) Mean radial peak-position error as a function of the resized image size.
    Error bars indicate the standard deviation across images.}
    \label{fig:resolution_convergence_peak_position}
\end{figure*}

\begin{table*}[t]
\centering
\small
\caption{\textbf{Material inventory and per-fold dataset roles
in the eight-fold leave-one-material-out evaluation.}
The experimental material set is
$\mathcal{M}_{\mathrm{exp}}
=\{\mathrm{Al,Cu,Fe,Mo,Nb,Ni,Ti,Zn}\}$.
Each fold holds out one material $m$ from this set,
reserving all its simulated and experimental scans for
outer testing. The remaining data in each domain are
partitioned into training, validation, and internal test
subsets using an 80\%/10\%/10\% split.
Experimental training contains 56 labeled anchors
(10\% of the training subset) and 504 unlabeled scans
per fold. Validation data are used for early stopping
and model selection; validation and test scans are
excluded from anchor selection.
The Test/Eval.\ column reports internal test counts
for development rows and independent evaluation counts
for outer-test and OOD rows.
SiC and noise-only images belong to a separate
uncertainty evaluation.}
\label{tab:dataset_fix}
\renewcommand{\arraystretch}{1.15}
\begin{tabular*}{\textwidth}
{@{\extracolsep{\fill}}llp{0.37\textwidth}cccc}
\hline
\textbf{Dataset}
& \textbf{Domain}
& \textbf{Materials}
& \textbf{Total}
& \textbf{Train}
& \textbf{Val.}
& \textbf{Test/Eval.} \\
\hline

Known-class inventory
& Simulation
& Ag, Al, Be, Cu, Fe, Hf, Mg, Mo, Nb, Ni, Pb, Pt,
Sc, Ta, Ti, V, Zn, Zr
& 900 & -- & -- & -- \\

Known-class inventory
& Experiment
& Al, Cu, Fe ($\alpha$-iron), Mo, Nb, Ni, Ti, Zn
& 800 & -- & -- & -- \\

\hline

Per-fold development
& Simulation
& Listed simulated materials excluding $m$
& 850 & 680 & 85 & 85 \\

Per-fold outer test
& Simulation
& $m$: Al, Cu, Fe, Mo, Nb, Ni, Ti, or Zn
& 50 & 0 & 0 & 50 \\

\hline

Per-fold development
& Experiment
& Al, Cu, Fe, Mo, Nb, Ni, Ti, Zn, excluding $m$
& 700 & 560 & 70 & 70 \\

Per-fold outer test
& Experiment
& $m$: Al, Cu, Fe, Mo, Nb, Ni, Ti, or Zn
& 100 & 0 & 0 & 100 \\

\hline

OOD evaluation
& Experiment
& Single-crystal and polycrystalline SiC
& 100 & 0 & 0 & 100 \\

OOD evaluation
& Experiment
& Noise-only images
& 100 & 0 & 0 & 100 \\

\hline
\end{tabular*}
\end{table*}

\begin{table*}[t]
\centering
\small
\caption{\textbf{Dataset partitions used in the main
domain-incremental sequential-update evaluation.}
Block identities remain fixed when their introduction order
changes. Image-level training, validation, and test roles are
preserved. Experimental blocks $D_1$--$D_3$ share Al, Mo, Ti,
and Zn, with Fe, Nb, and Ni occurring in $D_1$, $D_2$, and
$D_3$, respectively; experimental $D_4$ contains FCC Cu only.
Every simulated block contains all three crystal classes.
Recurring materials
use distinct image identities but may include correlated
measurements of the same specimen.}
\label{tab:dataset_summary}
\renewcommand{\arraystretch}{1.15}
\begin{tabular*}{\textwidth}
{@{\extracolsep{\fill}}llp{0.34\textwidth}cccc}
\hline
\textbf{Partition}
& \textbf{Domain}
& \textbf{Materials}
& \textbf{Total}
& \textbf{Train}
& \textbf{Val.}
& \textbf{Test} \\
\hline

$\mathcal{D}^{(1)}_{\mathrm{sim}}$
& Simulation
& Al, Be, Cu, Fe, Hf, Mg, Mo, Pb, Pt, Sc, Ta, Zn, Zr
& 247 & 205 & 20 & 22 \\

$\mathcal{D}^{(1)}_{\mathrm{exp}}$
& Experiment
& Al, Fe, Mo, Ti, Zn
& 233 & 186 & 22 & 25 \\

\hline

$\mathcal{D}^{(2)}_{\mathrm{sim}}$
& Simulation
& Al, Be, Cu, Hf, Mg, Mo, Nb, Pb, Pt, Sc, Ta, Zn, Zr
& 252 & 200 & 26 & 26 \\

$\mathcal{D}^{(2)}_{\mathrm{exp}}$
& Experiment
& Al, Mo, Nb, Ti, Zn
& 233 & 192 & 22 & 19 \\

\hline

$\mathcal{D}^{(3)}_{\mathrm{sim}}$
& Simulation
& Al, Be, Cu, Hf, Mg, Mo, Ni, Pb, Pt, Sc, Ta, Zn, Zr
& 251 & 197 & 28 & 26 \\

$\mathcal{D}^{(3)}_{\mathrm{exp}}$
& Experiment
& Al, Mo, Ni, Ti, Zn
& 234 & 184 & 25 & 25 \\

\hline

$\mathcal{D}^{(4)}_{\mathrm{sim}}$
& Simulation
& Ag, Ti, V
& 150 & 120 & 15 & 15 \\

$\mathcal{D}^{(4)}_{\mathrm{exp}}$
& Experiment
& Cu
& 100 & 80 & 10 & 10 \\

\hline
\end{tabular*}
\end{table*}

\begin{table}[htbp]
\small
\caption{\textbf{Record numbers for ICSD CIF files, accessed from FIZ Karlsruhe on September 12, 2025}}
\label{tab:ICSD_reference}
\begin{tabular*}{\columnwidth}{@{\extracolsep{\fill}}ll}
\hline
\textbf{Element} & \textbf{ICSD record number} \\
\hline
Ag & ICSD 64706\\
Al & ICSD 64700\\
Cu & ICSD 136042\\
Ni & ICSD 37502\\
Pb & ICSD 96501\\
Pt & ICSD 64919\\
Mo & ICSD 76279\\
Nb & ICSD 76011\\
Fe & ICSD 149920\\
Ta & ICSD 76014\\
V & ICSD 171003\\
Zr & ICSD 52544\\
Be & ICSD 1425\\
Hf & ICSD 76142\\
Mg & ICSD 76748\\
Sc & ICSD 164088\\
Ti & ICSD 43614\\
Zn & ICSD 64990\\

\hline
\end{tabular*}
\end{table}

This work developed an uncertainty-aware, continually updatable
analysis module for autonomous materials characterization
workflows. The general objective is to support reliable
interpretation of sequentially acquired, high-dimensional
characterization data while addressing practical challenges
associated with deployment in evolving experimental environments,
including limited labeled data, distribution mismatch across data
sources, overconfident predictions on unfamiliar inputs, and
degradation of previously learned knowledge during model updates.
The framework is demonstrated through
sequential interpretation of 2D XRD patterns, where the specific
task is to classify crystal structures from simulated and
experimental diffraction data.

The proposed workflow integrates three main components.
First, a CNN--SNGP model was used to perform crystal-structure
classification while providing distance-aware uncertainty
estimates. Second, simulation-to-experiment domain adaptation
was incorporated to improve transfer from labeled simulated
XRD patterns to partially labeled experimental measurements.
Third, a continual-learning strategy was introduced to enable
incremental model updating as new data blocks become available,
without requiring full retraining on all accumulated data.
Together, these components form a modular framework for
uncertainty-aware and adaptive XRD analysis from sequential
simulated and experimental data blocks.

The results provide several major observations.
The CNN--SNGP model improved the organization of the latent
feature space and provided better
predictive-probability quality than the conventional CNN
baseline, with unfamiliar-input detection
evaluated separately using noise-only and SiC inputs.
Because SNGP operates on learned feature representations,
it can also be paired with suitable encoders for other
high-dimensional materials inputs, including spectra,
microscopy images, and atomic-structure representations.
Domain adaptation reduced the discrepancy between simulated
and experimental feature distributions, improving the ability
of simulation-trained models to support experimental prediction.
Repeated leave-one-material-out evaluation
further demonstrated improved average experimental performance
with the adapted model, while revealing remaining challenges
in generalization to unfamiliar materials.
Across multiple random seeds and data-block
orders, the continual-learning strategy maintained strong
classification performance while improving retention of
previously learned knowledge compared with naive fine-tuning.
It also improved predictive-probability quality
relative to naive fine-tuning and reduced training time relative
to joint retraining under the same backbone, although it did
not uniformly outperform simpler replay controls.
Experimental-label and replay-memory sensitivity analyses
further characterized the dependence on supervision and
retained historical data.
These findings demonstrate that uncertainty estimation,
domain alignment, and continual model updating provide
complementary benefits for adaptive materials characterization.

This study demonstrates a proof-of-concept workflow using
single-phase, high-symmetry BCC, FCC, and HCP materials as a
representative classification task.
The evaluation uses predefined data blocks
and does not demonstrate closed-loop experimental operation.
The broader significance of the framework is that it provides
a pathway toward characterization models that can be updated
as live-stream experimental data become available, while
retaining uncertainty information needed for reliable
downstream decision-making. Such capabilities are important
for autonomous characterization systems, where models must
operate under changing data availability, evolving sample
sets, and incomplete prior knowledge.

Several limitations remain. The current evaluation is
restricted to a limited number of simulated and experimental
2D XRD datasets and to crystal-structure-level classification
rather than full phase identification or quantitative
structural refinement. Future work will extend the framework
to multiphase materials, lower-symmetry structures, amorphous
systems, stronger texture effects, and changing instrument
conditions. Additional directions include predicting lattice
parameters, orientation, strain-related features, and phase
fractions directly from 2D diffraction images, as well as
incorporating uncertainty-based decision rules into closed-loop
autonomous experiments.
For materials discovery, such integration
could connect XRD-derived structural information with
composition--property models to guide the selection of new
compositions for synthesis and evaluation, with prospective
validation needed to establish improvements in discovery
efficiency.

\section*{Appendix}
\appendix

\section{Resolution-convergence analysis for XRD image resizing}
\label{app:resolution_convergence}

To justify the resizing of the original $1024 \times 1024$ detector images to
$256 \times 256$ pixels, we performed a resolution-convergence analysis based on
diffraction-ring position preservation. The purpose of this analysis was not to
preserve every detector-scale pixel intensity, but to verify that the main
diffraction-ring features relevant to crystal-structure classification were
retained after resizing.

Candidate resolutions of $512 \times 512$, $384 \times 384$, $256 \times 256$,
$128 \times 128$, and $64 \times 64$ pixels were evaluated. For each candidate
resolution, the original $1024 \times 1024$ image was first downsampled to the
target size using bilinear interpolation and then upsampled back to
$1024 \times 1024$ pixels for direct comparison. Representative resized-restored
images are shown in Figure~\ref{fig:resolution_convergence_peak_position}a.

To quantify this behavior, a radial intensity profile was computed for each
original and resized-restored image by averaging the image intensity as a
function of radial distance from the detector center. The main radial peaks were
then identified from the radial profiles, corresponding to the dominant
diffraction-ring positions in the 2D XRD images. For each candidate resolution,
the mean radial peak-position error was calculated as the average absolute
difference between the main peak positions in the original and resized-restored
radial profiles. As shown in Figure~\ref{fig:resolution_convergence_peak_position}b, the mean radial
peak-position error remains small for $512 \times 512$, $384 \times 384$, and
$256 \times 256$ images, indicating that the main diffraction-ring locations are
largely preserved after resizing. In contrast, further reduction to
$128 \times 128$ and $64 \times 64$ results in a larger increase in peak-position
error and greater variability across images. Therefore, $256 \times 256$ pixels
was selected as a practical compromise between preserving the diffraction-ring
information needed for crystal-structure classification and reducing computational cost during model training.

\section{Supplementary Dataset Details}
\label{app:dataset}

Tables~\ref{tab:dataset_fix} and~\ref{tab:dataset_summary}
summarize the dataset compositions and train--test partitions
used in the fixed-training and sequential-update evaluations,
respectively. The ICSD record numbers for the crystallographic structures used
to generate the simulated XRD patterns are listed in
Table~\ref{tab:ICSD_reference}. A more detailed dataset inventory, including
material identifiers, exposure times, sequential-block assignments,
training/validation/test counts, anchor budgets, and evaluation
roles, is provided in Table~\ref{tab:dataset_inventory_current_R1}.

\makeatletter
\newif\ifDatasetInventoryRestoreTwoColumn
\if@twocolumn
  \DatasetInventoryRestoreTwoColumntrue
  \onecolumn
\else
  \DatasetInventoryRestoreTwoColumnfalse
  \clearpage
\fi
\makeatother

\fontsize{8.5}{10}\selectfont
\setlength{\tabcolsep}{3pt}
\renewcommand{\arraystretch}{1.1}
\newlength{\DatasetInventoryWidth}
\setlength{\DatasetInventoryWidth}{\dimexpr\linewidth-12\tabcolsep\relax}
\setlength{\LTleft}{0pt}
\setlength{\LTright}{0pt}
\setlength{\LTcapwidth}{\linewidth}
\begin{longtable}{@{}>{\raggedright\arraybackslash}p{0.070\DatasetInventoryWidth}>{\raggedright\arraybackslash}p{0.150\DatasetInventoryWidth}>{\raggedright\arraybackslash}p{0.090\DatasetInventoryWidth}>{\raggedright\arraybackslash}p{0.055\DatasetInventoryWidth}>{\raggedright\arraybackslash}p{0.230\DatasetInventoryWidth}>{\raggedright\arraybackslash}p{0.150\DatasetInventoryWidth}>{\raggedright\arraybackslash}p{0.255\DatasetInventoryWidth}@{}}
\caption{Dataset composition and evaluation assignments.
CL denotes continual learning; Tr/V/Te gives training/validation/test
image counts. The experimental anchors/hidden entries identify the
requested overall training-label budget of 10\%; they do not give
realized counts for each row or block. The remaining training labels
are hidden. The fixed and grouped comparisons also have a requested
10\% experimental training-label budget.
L denotes the fixed
$D^{(1)}$--$D^{(3)}$ comparison, and G denotes the eight-material held-out
evaluation with three repeats. ``Held out / development'' means that a
material is the outer test set in its own holdout fold and development
data in other folds. Exp. gives exposure time; N/A denotes not applicable.}
\label{tab:dataset_inventory_current_R1}\\
\toprule
Type & Material & Structure & \shortstack[l]{Exp.\\(s)} & CL block: Tr/V/Te & CL anchors / hidden & Fixed / grouped roles \\
\midrule\endfirsthead
\multicolumn{7}{l}{\tablename\ \thetable\ (continued)}\\
\toprule
Type & Material & Structure & \shortstack[l]{Exp.\\(s)} & CL block: Tr/V/Te & CL anchors / hidden & Fixed / grouped roles \\
\midrule\endhead
\midrule\multicolumn{7}{r}{Continued on next page}\\\endfoot
\bottomrule\endlastfoot
Sim. & Ag & FCC & N/A & D4: 42/4/4 & All training labels & L: full material holdout (50)\newline G: development only \\
Sim. & Al & FCC & N/A & D1: 14/2/1\newline D2: 13/3/0\newline D3: 14/2/1 & All training labels & L Tr/V/Te: 41/7/2\newline G: held out / development \\
Sim. & Be & HCP & N/A & D1: 13/1/2\newline D2: 13/2/2\newline D3: 13/2/2 & All training labels & L Tr/V/Te: 39/5/6\newline G: development only \\
Sim. & Cu & FCC & N/A & D1: 14/1/1\newline D2: 15/1/1\newline D3: 14/1/2 & All training labels & L Tr/V/Te: 43/3/4\newline G: held out / development \\
Sim. & Fe & BCC & N/A & D1: 43/1/6 & All training labels & L Tr/V/Te: 43/1/6\newline G: held out / development \\
Sim. & Hf & HCP & N/A & D1: 13/3/1\newline D2: 13/2/2\newline D3: 12/3/1 & All training labels & L Tr/V/Te: 38/8/4\newline G: development only \\
Sim. & Mg & HCP & N/A & D1: 13/2/1\newline D2: 14/1/2\newline D3: 13/2/2 & All training labels & L Tr/V/Te: 40/5/5\newline G: development only \\
Sim. & Mo & BCC & N/A & D1: 13/2/1\newline D2: 13/2/2\newline D3: 13/2/2 & All training labels & L Tr/V/Te: 39/6/5\newline G: held out / development \\
Sim. & Nb & BCC & N/A & D2: 38/6/6 & All training labels & L Tr/V/Te: 38/6/6\newline G: held out / development \\
Sim. & Ni & FCC & N/A & D3: 37/6/7 & All training labels & L Tr/V/Te: 37/6/7\newline G: held out / development \\
Sim. & Pb & FCC & N/A & D1: 13/1/2\newline D2: 13/2/2\newline D3: 13/2/2 & All training labels & L Tr/V/Te: 39/5/6\newline G: development only \\
Sim. & Pt & FCC & N/A & D1: 13/1/2\newline D2: 13/1/3\newline D3: 13/1/3 & All training labels & L Tr/V/Te: 39/3/8\newline G: development only \\
Sim. & Sc & HCP & N/A & D1: 14/2/1\newline D2: 14/1/2\newline D3: 14/1/1 & All training labels & L Tr/V/Te: 42/4/4\newline G: development only \\
Sim. & Ta & BCC & N/A & D1: 14/2/0\newline D2: 14/3/0\newline D3: 14/3/0 & All training labels & L Tr/V/Te: 42/8/0\newline G: development only \\
Sim. & Ti & HCP & N/A & D4: 38/4/8 & All training labels & L: full material holdout (50)\newline G: held out / development \\
Sim. & V & BCC & N/A & D4: 40/7/3 & All training labels & L: full material holdout (50)\newline G: development only \\
Sim. & Zn & HCP & N/A & D1: 14/1/2\newline D2: 14/1/1\newline D3: 14/2/1 & All training labels & L Tr/V/Te: 42/4/4\newline G: held out / development \\
Sim. & Zr & BCC & N/A & D1: 14/1/2\newline D2: 13/1/3\newline D3: 13/1/2 & All training labels & L Tr/V/Te: 40/3/7\newline G: development only \\
Exp. & Mo & BCC & 1 & D1: 13/2/2\newline D2: 16/1/0\newline D3: 13/3/0 & 10\% & L Tr/V/Te: 42/6/2\newline G: held out / development \\
Exp. & Mo & BCC & 3 & D1: 15/1/1\newline D2: 11/3/2\newline D3: 14/1/2 & 10\% & L Tr/V/Te: 40/5/5\newline G: held out / development \\
Exp. & Fe & BCC & 1 & D1: 42/3/5 & 10\% & L Tr/V/Te: 42/3/5\newline G: held out / development \\
Exp. & Fe & BCC & 3 & D1: 39/7/4 & 10\% & L Tr/V/Te: 39/7/4\newline G: held out / development \\
Exp. & Zn & HCP & 1 & D1: 11/2/3\newline D2: 13/2/2\newline D3: 14/1/2 & 10\% & L Tr/V/Te: 38/5/7\newline G: held out / development \\
Exp. & Zn & HCP & 3 & D1: 15/1/1\newline D2: 14/1/2\newline D3: 12/2/2 & 10\% & L Tr/V/Te: 41/4/5\newline G: held out / development \\
Exp. & Ti & HCP & 1 & D1: 16/0/3\newline D2: 9/3/2\newline D3: 14/1/2 & 10\% & L Tr/V/Te: 39/4/7\newline G: held out / development \\
Exp. & Ti & HCP & 3 & D1: 10/3/1\newline D2: 17/0/2\newline D3: 13/2/2 & 10\% & L Tr/V/Te: 40/5/5\newline G: held out / development \\
Exp. & Nb & BCC & 1 & D2: 44/5/1 & 10\% & L Tr/V/Te: 44/5/1\newline G: held out / development \\
Exp. & Nb & BCC & 3 & D2: 43/4/3 & 10\% & L Tr/V/Te: 43/4/3\newline G: held out / development \\
Exp. & Cu & FCC & 1 & D4: 34/10/6 & 10\% & L: full material holdout (50)\newline G: held out / development \\
Exp. & Cu & FCC & 3 & D4: 46/0/4 & 10\% & L: full material holdout (50)\newline G: held out / development \\
Exp. & Al & FCC & 1 & D1: 11/1/2\newline D2: 14/2/2\newline D3: 12/2/4 & 10\% & L Tr/V/Te: 37/5/8\newline G: held out / development \\
Exp. & Al & FCC & 3 & D1: 14/2/3\newline D2: 11/1/3\newline D3: 13/2/1 & 10\% & L Tr/V/Te: 38/5/7\newline G: held out / development \\
Exp. & Ni & FCC & 1 & D3: 38/7/5 & 10\% & L Tr/V/Te: 38/7/5\newline G: held out / development \\
Exp. & Ni & FCC & 3 & D3: 41/4/5 & 10\% & L Tr/V/Te: 41/4/5\newline G: held out / development \\
Exp. OOD & Noise only & N/A & 1 & Excluded & None & Excluded from all classification partitions \\
Exp. OOD & Noise only & N/A & 3 & Excluded & None & Excluded from all classification partitions \\
Exp. OOD & SiC & 4H & 1 & Excluded & None & Excluded from all classification partitions \\
Exp. OOD & SiC & 4H & 3 & Excluded & None & Excluded from all classification partitions \\
\end{longtable}

\ifDatasetInventoryRestoreTwoColumn\twocolumn\fi

\section{Sensitivity to the Experimental Training-Anchor Budget}
\label{app:anchor_budget_sensitivity}

We evaluated CNN--SNGP--adapt using requested experimental
training-anchor fractions of 0\%, 1\%, 5\%, 10\%, 20\%, 30\%,
50\%, and 100\%, with three initializations per configuration.
The same labeled experimental validation set was retained
at every budget. Validation and test scans were excluded
from anchor selection.

\begin{figure}[t]
\centering
\includegraphics[width=\linewidth]{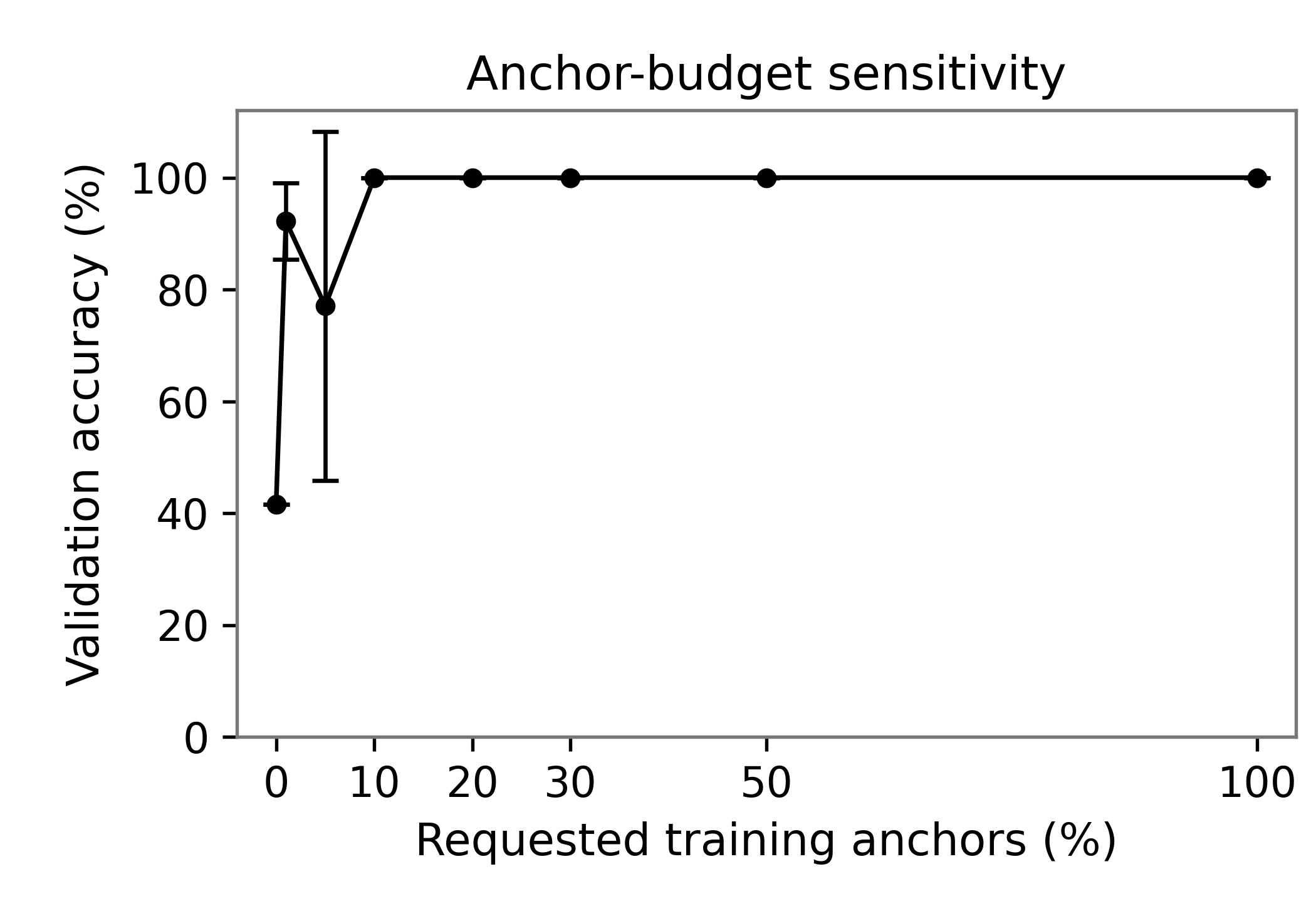}
\caption{
Experimental validation accuracy of CNN--SNGP--adapt versus
the requested training-anchor fraction. Markers and error bars
show the mean and sample standard deviation across three
initializations. All configurations retain the same 77 labeled
experimental validation scans.}
\label{fig:anchor_budget_sensitivity}
\end{figure}

As shown in Figure~\ref{fig:anchor_budget_sensitivity},
validation accuracy reached $100.00 \pm 0.00\%$ at 10\%
and remained unchanged at all larger tested budgets.
We therefore selected 10\% as the smallest tested fraction
attaining this validation-accuracy plateau. Selection was
based on validation accuracy.

\section{Sensitivity to Replay-Memory Capacity}
\label{app:replay_memory_sensitivity}

We evaluate the sensitivity of proposed continual learning
to replay-memory capacity using the reference block order
$D_1$--$D_2$--$D_3$--$D_4$ and three random seeds.
Table~\ref{tab:continual_replay_analysis} reports final-stage
experimental accuracy pooled across test scans from all
four blocks.

Mean experimental accuracy increases with replay capacity,
although variability across seeds does not decrease
consistently. At zero capacity, both replay supervision
and replay-based distillation are disabled. This setting
therefore differs from the memory-free LwF-style control,
which applies distillation to current inputs.
The results characterize sensitivity within the tested
capacities and reference block order.

\begin{table*}[t]
\centering
\small
\setlength{\tabcolsep}{3pt}
\caption{\textbf{Sensitivity to replay-memory capacity.}
Proposed continual learning is evaluated using the reference
order $D_1$--$D_2$--$D_3$--$D_4$.
Values are mean $\pm$ sample standard deviation across
three random seeds. Exp. Acc.\ denotes final-stage
experimental accuracy pooled across test scans from all
four blocks. These reference-order results are reported
separately from the order-averaged primary comparison.
Zero capacity disables replay supervision and replay-based
distillation.}
\label{tab:continual_replay_analysis}

\begin{tabular*}{\textwidth}
{@{\extracolsep{\fill}}lccccc}
\hline
\shortstack{\textbf{Replay capacity}\\\textbf{(images)}}
& \textbf{0}
& \textbf{32}
& \textbf{64}
& \textbf{128}
& \textbf{256} \\
\hline
Exp. Acc. (\%)
& $88.19 \pm 0.73$
& $94.94 \pm 7.70$
& $95.78 \pm 7.31$
& $97.05 \pm 1.46$
& $97.89 \pm 0.73$ \\
\hline
\end{tabular*}
\end{table*}

\section{Supplementary comparison of CNN--DKL and CNN--SNGP-adapt}
\label{app:supp_cnn_dkl}

We compared CNN--DKL \cite{wilson2016deep} with CNN--SNGP-adapt
using identical $D^{(1)}$--$D^{(3)}$ training and test partitions,
with experimental Cu reserved as a material-held-out evaluation
set. This supplementary comparison uses a single dataset
partition rather than the repeated eight-material-held-out
protocol used in the main evaluation.
Both models achieved 100\% classification accuracy on the
simulated and experimental training and test subsets of
$D^{(1)}$--$D^{(3)}$.
Table~\ref{tab:supp_cnn_dkl} additionally reports performance
on the separate simulated holdout and experimental Cu datasets
from $D^{(4)}$.

\begin{table}[htbp]
\centering
\small
\setlength{\tabcolsep}{4pt}
\caption{Classification accuracy (\%) of CNN--DKL and
CNN--SNGP-adapt. CNN denotes convolutional neural network,
DKL denotes deep kernel learning, and SNGP denotes
spectral-normalized neural Gaussian process.
The suffix ``adapt'' indicates domain adaptation.
Sim. and Exp. denote simulated and experimental data.
$D^{(4)}$ is evaluated separately from $D^{(1)}$--$D^{(3)}$.}
\label{tab:supp_cnn_dkl}
\begin{tabular}{@{}lcc@{}}
\toprule
Evaluation set & CNN--DKL & \shortstack{CNN--SNGP\\-adapt} \\
\midrule
$D^{(1)}$--$D^{(3)}$, Sim. train & 100.00 & 100.00 \\
$D^{(1)}$--$D^{(3)}$, Exp. train & 100.00 & 100.00 \\
$D^{(1)}$--$D^{(3)}$, Sim. test  & 100.00 & 100.00 \\
$D^{(1)}$--$D^{(3)}$, Exp. test  & 100.00 & 100.00 \\
\midrule
$D^{(4)}$, Sim. holdout    & 98.00 & 98.00 \\
$D^{(4)}$, Exp. Cu holdout & 43.00 & 61.00 \\
\bottomrule
\end{tabular}
\end{table}

\section*{Author contributions}

Zihan Wang: Conceptualization, Methodology, Validation, Formal analysis, Investigation, Data curation, Visualization, and Writing -- original draft. Hyun Sang Park: Investigation, Data curation, Software, Writing -- original draft, and Writing -- review \& editing. Ali Rachidi: Software, Resources, Data curation, and Writing -- review \& editing. David Elbert: Software, Resources, Supervision, and Writing -- review \& editing. Todd Hufnagel: Conceptualization, Resources, Supervision, and Writing -- review \& editing. Wei Chen: Conceptualization, Methodology, Supervision, Project administration, Funding acquisition, and Writing -- review \& editing.

\section*{Conflicts of interest}

\section*{Data availability}

The data supporting this study, including the simulated and experimental 2D XRD datasets, is available at \url{https://github.com/zihanw97/DL_XRD_R1/releases/tag/r1-inputs-20261109}. The custom code is available at \url{https://github.com/zihanw97/DL_XRD_R1}.

\section*{Acknowledgements}

This collaboration is supported by the U.S. Army Combat Capabilities Development Command Army Research Laboratory under Cooperative Agreement Number W911NF-22-0121. Zihan Wang and Wei Chen also received partial support from the Air Force Office of Scientific Research under award number FA9550-24-1-0301.

The authors used AI tools to assist with rephrasing text and improving the language and clarity of the manuscript. The authors reviewed the resulting text and take full responsibility for the final content.



\balance

\renewcommand\refname{References}

\bibliography{rsc} 
\bibliographystyle{rsc} 
\end{document}